\documentclass[letterpaper, 10 pt, conference]{ieeeconf}  

\IEEEoverridecommandlockouts                              

\usepackage{amsmath}
\usepackage{amssymb}
\usepackage{wrapfig}
\usepackage{subfig}
\usepackage{caption}
\usepackage{color}
\usepackage{algorithm}
\usepackage{algorithmic}
\usepackage{graphicx}

\usepackage{pgfplots}
\usepackage{tikz}
\usepackage{tikz-3dplot}

\usepackage{ragged2e}
\usepackage{fp}

\usepackage{ifthen}

\usepackage{tikz}
\usetikzlibrary{shapes.geometric, arrows, calc,automata,positioning,decorations,decorations.text}
\usepackage{multirow}

\usepackage{algorithm}
\usepackage{algorithmic}
\usepackage{graphicx}
\usepackage[english]{babel}
\usepackage{array}
\usepackage[export]{adjustbox}

\graphicspath{{images/}}

\usepackage{xcolor}

\title{\LARGE \bf
Real Time Incremental Foveal Texture Mapping for Autonomous Vehicles
}

\author{Ashish~Kumar$^{1}$, James R. McBride$^{2}$, Gaurav~Pandey$^{2}$
\thanks{$^{1}$Mr. Ashish kumar is with the Department
of Electrical Engineering, Indian Institute of Technology, Kanpur
        {\tt\small krashish@iitk.ac.in}}%
\thanks{$^{2}$Dr. James R. McBride and Dr. Gaurav Pandey are with Research \& Innovation Center, Ford Motor Company
        {\tt\small [jmcbride, gpandey2]@ford.com}}%
}

\begin{document}

\maketitle
\thispagestyle{empty}
\pagestyle{empty}

\begin{justify}

\begin{abstract}
We propose an end-to-end real time framework to generate high resolution graphics grade textured 3D map of urban environment. The generated detailed map finds its application in the precise localization and navigation of autonomous vehicles. It can also serve as a virtual test bed for various vision and planning algorithms as well as a background map in the computer games.
In this paper, we focus on two important issues: (\textit{i}) incrementally generating a map with coherent 3D surface, in real time and (\textit{ii}) preserving the quality of color texture. To handle the above issues, firstly, we perform a \texttt{pose-refinement} procedure which leverages camera image information, Delaunay triangulation and existing scan matching techniques to produce high resolution 3D map from the sparse input LIDAR scan. This 3D map is then texturized and accumulated by using a novel technique of \texttt{ray-filtering} which handles occlusion and inconsistencies in \texttt{pose-refinement}. Further, inspired by human fovea, we introduce \texttt{foveal-processing} which significantly reduces the computation time and also assists \texttt{ray-filtering} to maintain consistency in color texture and coherency in 3D surface of the output map. Moreover, we also introduce texture~error~(TE) and mean~texture~mapping~error~(MTME), which provides quantitative measure of texturing and overall quality of the textured maps. 

\end{abstract}

\end{justify}

\section{Introduction}
Recent advances in the sensor technology (\cite{velodyne-wp,ladybug3}) and a breakthrough in real world algorithmic visual inference (\cite{chabot2017deep,chen2016monocular,fasterrcnn,maskrcnn,pspnet}) has brought the fully autonomous navigation/driving one step closer to the reality, perhaps most visible in (\cite{darpa-2009, ford, tesla}). In this work, we focus on urban maps for localization of an autonomous navigation system (ANS). 
\par
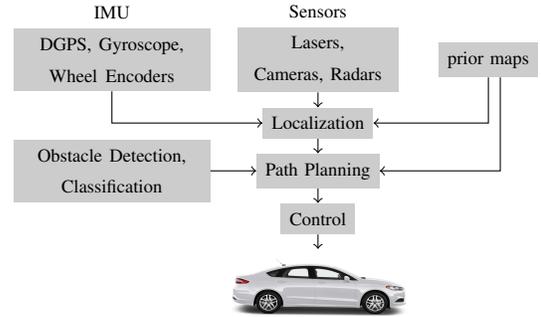
\begin{figure}[t]
\centering

\begin{tikzpicture}

\node (sensors) [rectangle, fill=white!80!black, minimum width=1ex, minimum height =1ex,text width=12ex,align=center, label={[]above:\scriptsize Sensors}]{\scriptsize Lasers, Cameras, Radars};

\node (navigation) [rectangle, left of=sensors, fill=white!80!black, minimum width=1ex, minimum height =1ex,align=center,xshift=-11ex,yshift=0.0ex,text width=15ex,label={[]above:\scriptsize IMU}]{\scriptsize DGPS, Gyroscope, \\
Wheel Encoders};

\node (priormaps) [rectangle, right of=sensors,  fill=white!80!black, minimum width=1ex, minimum height =1ex,xshift=8ex,yshift=0ex]{\scriptsize prior maps};

\node (localization) [rectangle,below of=sensors, fill=white!80!black, minimum width=1ex, minimum height =1ex,yshift=1ex]{\scriptsize Localization};

\node (planning) [rectangle, below of=localization, fill=white!80!black, minimum width=1ex, minimum height =1ex,yshift=2.3ex]{\scriptsize Path Planning};

\node (control) [rectangle, below of=planning, fill=white!80!black, minimum width=1ex, minimum height =1ex,yshift=2.3ex]{\scriptsize Control};

\node (cv) [rectangle, left of=planning, fill=white!80!black, minimum width=1ex, minimum height =1ex,align=center,xshift=-11ex,yshift=-0ex,text width=15ex]{\scriptsize Obstacle Detection, Classification};

\node (car) [rectangle, below of=control, minimum width=1ex, minimum height =1ex,yshift=0.5ex,xshift=0ex,text width=15ex]{\includegraphics[width=15ex,height=5ex]{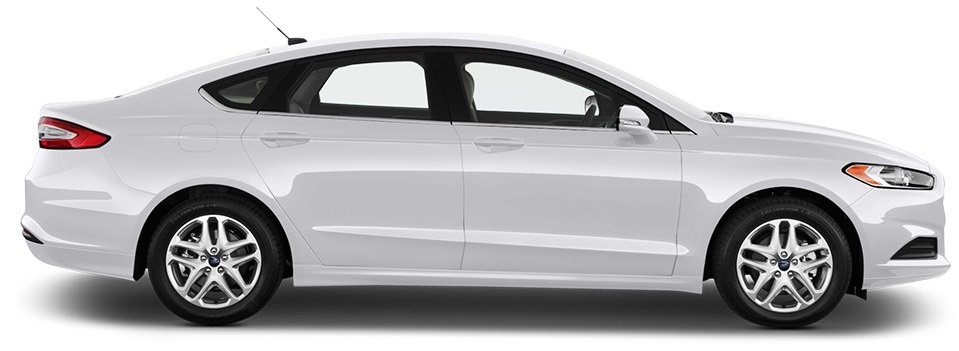}};

\draw [->] (navigation) |- (localization);
\draw [->] (sensors) -- (localization);
\draw [->] (priormaps) |- (localization);

\draw [->] (localization) -- (planning);
\draw [->] ($(priormaps.south) + (1ex,0)$) |- (planning);

\draw [->] (cv) -- (planning);
\draw [->] (planning) -- (control);
\draw [->] (control) --  (car.north);

\end{tikzpicture}

\caption{Autonomous Navigation System}
\label{fig_ans}
\end{figure}

Autonomous navigation system is fairly complex and consists of several modules as shown in fig. \ref{fig_ans}. In an ANS, localization (\cite{rwolcott15a,Levinson2010a,Wu2013}) happens to be one of the critical tasks, as it provides vehicle's location in the navigating environment. Although, centimeter accurate Global Positioning System (GPS) and Inertial Navigation System (INS) (\cite{sukkarieh1999high,barshan1995inertial}) are available but they are often quite expensive. Moreover, these costly and high accuracy GPS technologies fail in urban environment (due to multi-path error) and in regions where the line-of-sight to the GPS satellites is blocked (e.g. tunnels, underpasses, tree canopies etc.), resulting into a GPS-denied environment. In these scenarios, a vehicle/robot registers the current sensor data (LIDAR/camera) with the map of the environment (\texttt{prior~maps}) to localize itself. Hence, it becomes crucial to acquire high quality textured 3D maps of the environment for a safe and robust operation of autonomous vehicles. 
\par
Typically, the textured maps are generated by manually driving a surveying vehicle endowed with LIDARs, GPS/INS and cameras through the environment/area to be mapped. In general, the LIDAR is a short range device which provides a sparse 3D measurements. Hence, in order to generate a dense map of the environment, two approaches exists: (\textit{i}) acquire the scans of the area from different physical locations and accumulate them into a local frame by a standard simultaneous localization and mapping (SLAM) algorithm (\cite{graph-slam,olson-sgd,reustice-2006b,Durrant,grisetti10titsmag}) or (\textit{ii}) exploit the camera images for dense reconstruction of the map using bundle adjustment \cite{romanoni2015incremental, engelhard2011real, triggs1999bundle}.
\par
Generating a detailed texture map of the environment is a challenging problem mainly due to sparsity in LIDAR scans and inconsistency in pose estimation due to complex trajectories which sometimes are hard to optimize using SLAM. Moreover, $occlusion$ of 3D points in between LIDAR and camera frame appears as a bottleneck for detailed texture mapping. It arises due to different physical mounting locations of LIDAR and camera. Hence, in this paper, we propose an end-to-end real time framework which performs multi modal sensor data fusion (LIDAR scans, images, navigation data) to generate highly accurate, detailed textured 3D maps of urban areas while simultaneously handling the problems of occlusion and pose inconsistency. The framework outperforms recent work \cite{romanoni2015incremental,romanoni2017mesh} and achieves map error below $5$cm (in some cases $1$cm). 
\par
\begin{figure}[t]
\centering
\subfloat[Before \texttt{pose-refinement}]
{
\includegraphics[width=0.47\linewidth,height=8ex,frame]{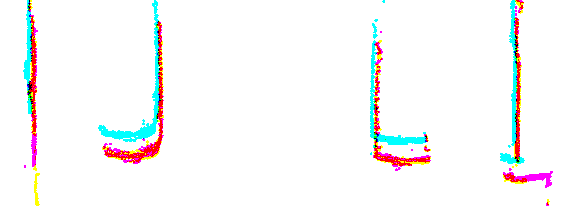}
}
\subfloat[After \texttt{pose-refinement}]
{
\includegraphics[width=0.47\linewidth,height=8ex,frame]{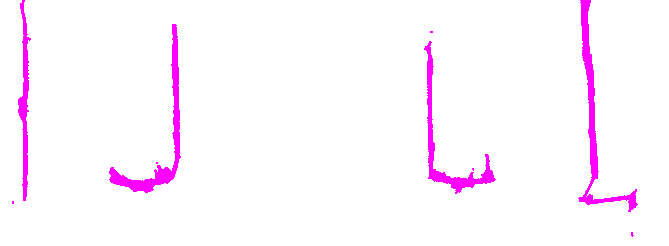}
}
\\
\subfloat[Foveal extraction]
{
\includegraphics[width=0.47\linewidth,height=8ex,frame]{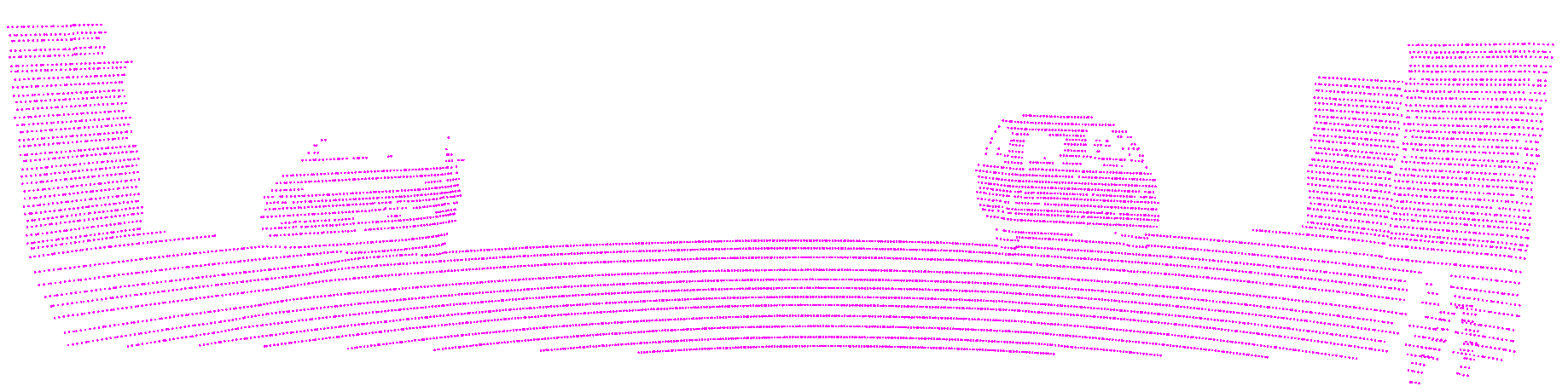}
}
\subfloat[Upsampling]
{
\includegraphics[width=0.47\linewidth,height=8ex,frame]{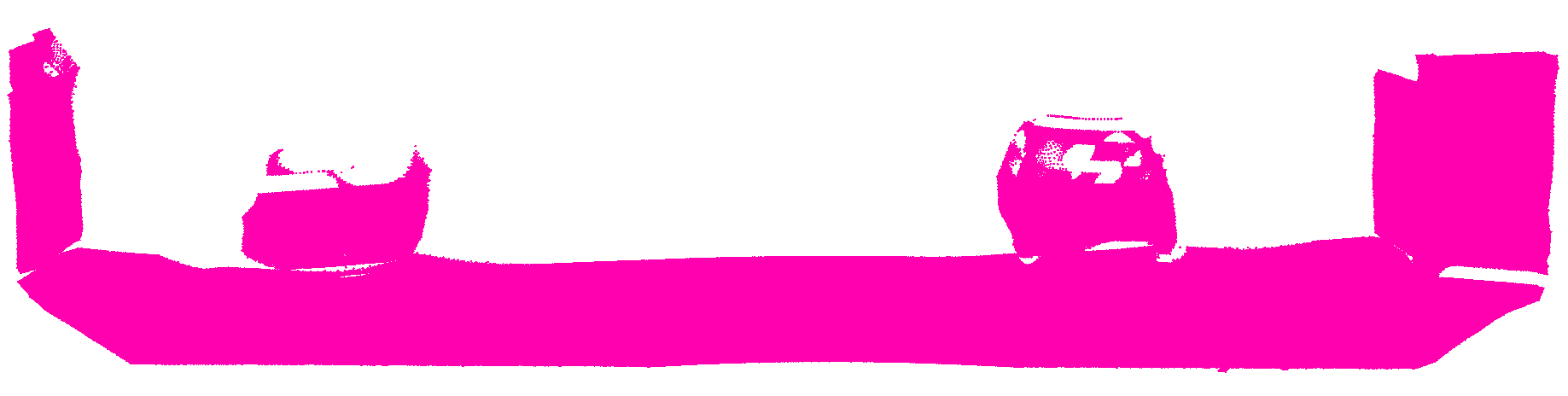}
}
\\
\subfloat[Foveal textured point cloud]
{
\includegraphics[width=0.47\linewidth,height=8ex]{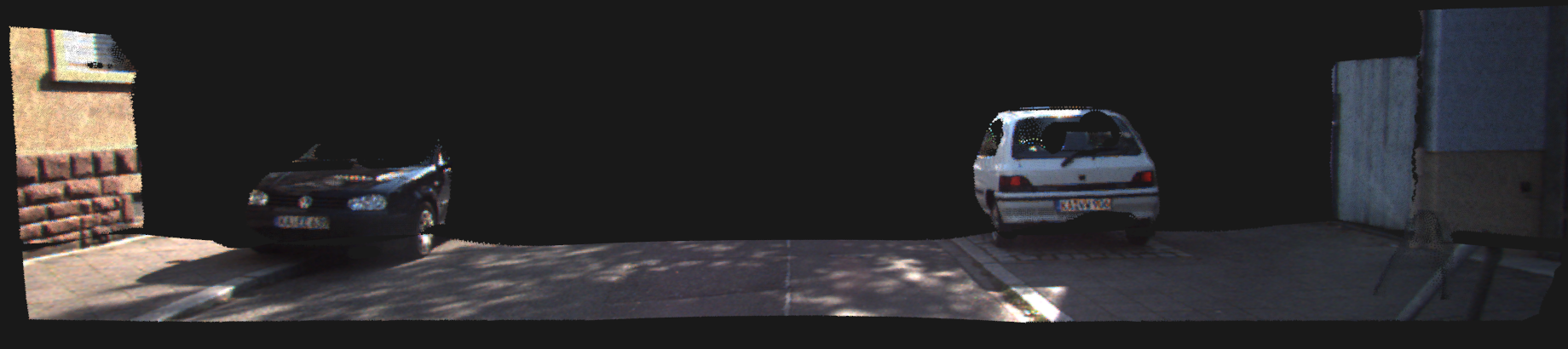}
\label{fig_foveal_extraction}
}
\subfloat[Corresponding ground truth image]
{
\includegraphics[width=0.47\linewidth,height=8ex]{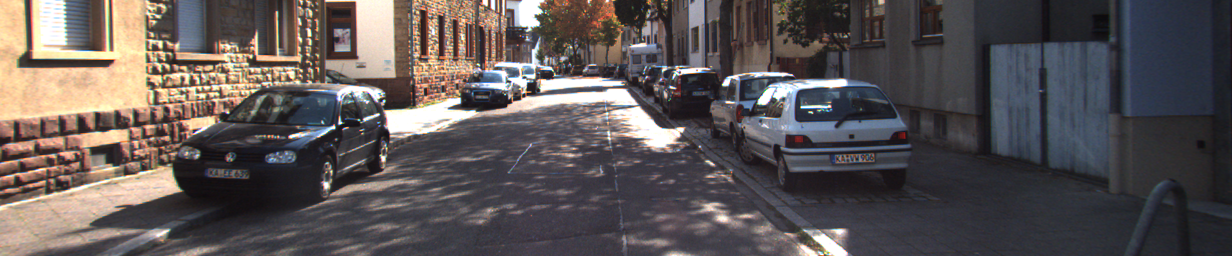}
}
\caption{Major components of the framework}
\label{fig_results_framework}
\end{figure}

The proposed framework (Fig. \ref{fig_results_framework}) consists of a novel \texttt{pose-refinement} (Sec. \ref{subsec_pose_refinement})
technique which facilitates high fidelity pairwise scan matching and produces accurately aligned dense version of input sparse scans. 
Once the dense 3D point cloud is obtained, we use a novel \texttt{ray-filtering} (Sec. \ref{subsec_ray_filtering}) technique to transfer color texture into the aligned scan and accumulate it in such a way that output map shows a high degree of coherent surface. This step only processes the \texttt{foveal~regions} (2D and 3D) defined by the novel \texttt{foveal-processing} concept which we have introduced in the sec. \ref{subsec_foveal} and is inspired from human fovea. This concept enhances the overall speed of the framework and improves the quality of the output map. Further, we introduce two new metrics to asses texture and overall quality of the map which are discussed in the sec. \ref{subsec_error_metric}. In the next section, we give an overview of the various topics involved in the design of this framework.

\section{Related Work}
A prevalent approach for the pose estimation is to use SLAM (\cite{graph-slam,olson-sgd,reustice-2006b,Durrant,grisetti10titsmag}), which requires all the data apriory. However, pose estimation can also be done by using scan matching techniques such as Standard-ICP \cite{besl1992method}, Point-to-Plane-ICP \cite{bergevin1996towards}, Generalized-ICP \cite{gicp}. The Standard-ICP minimizes point to point error (euclidean distances) while Point-to-Plane-ICP minimizes distance between a point and a locally estimated plane. GICP combines point-to-point and point-to-plane metric and achieves state-of-the-art results. In this paper, we choose to estimate the poses using scan matching techniques while focus remains on improving their accuracy. We do this due to two facts: (\textit{i}) scan matching performs better for better initial guess as well as for the lesser distances between scans to be aligned and (\textit{ii}) the state-of-the-art sensors (e.g. LIDAR, camera, INS/GPS) can provide data typically at the speed of $\sim10$fps. Due to the real time acquisition, LIDAR scans are separated by relatively short distances ($\sim0.5m$) and INS/GPS data can serve as a better initial guess for scan alignment.
\par
The existing scan matching algorithms are point hungry and upsampling often leads to improved scan matching performance. \cite{mls} uses local plane to estimate underlying surface using moving least squares and upsample the surrounding of the point uniformly. However, it does not perform in real time due to its exhaustive computationally intensive steps (Table \ref{tab_upsampling}). Whereas, the proposed upsampling technique performs in real time while approximately preserves the underlying surface.
\par
LIDAR data becomes unreliable as target distance increases and appears as major source of noise in the mapping process. To cop with this, we refer to the working of foveal vision in the human visual cortex. In this domain, the success of the attention based deep learning architectures for object detection \cite{gould2007peripheral,ba2014multiple,ablavatski2017enriched}, encourages us to employ foveal vision but we use it in an entirely different manner.
\par
The work in \cite{pan2009proforma} and \cite{litvinov2013incremental, lovi2011incremental,litvinov2014incremental}  proposes batch and incremental approaches respectively for texture mapping. The former performs the mapping by accounting all the viewing rays where as, the latter reconstruct the map incrementally by estimating boundary between free space and matter i.e. space carving. \cite{romanoni2015incremental} uses tracking of image edge points to incrementally reconstruct a less detailed map of the environment. Similar to us, \cite{romanoni2017mesh} jointly estimate a 3D map of the environment by using LIDAR scans and images. However, they follow batch reconstruction in contrast to the proposed work which is incremental in nature. Moreover, both \cite{romanoni2015incremental} and \cite{romanoni2017mesh} output a 3D mesh while the method proposed in this paper estimates a dense point cloud. \cite{romanoni2017mesh} discards moving objects such as pedestrian or bicycles. It also proposes a novel way to incrementally estimate the texture which is only evaluated qualitatively because in the previous literature, a quantitative metric to asses texture quality doesn't exist. In this work we propose such a metric and focus mainly on improving the overall map quality while leaving the moving object handling for future extension of the work.

\section{Methodology}
\label{methodology}
\subsection{Pose Refinement}
\label{subsec_pose_refinement}

\begin{figure}[t]
\centering
\vspace{2ex}
\subfloat[]
{
\includegraphics[width=0.15\linewidth,height=8ex,frame]{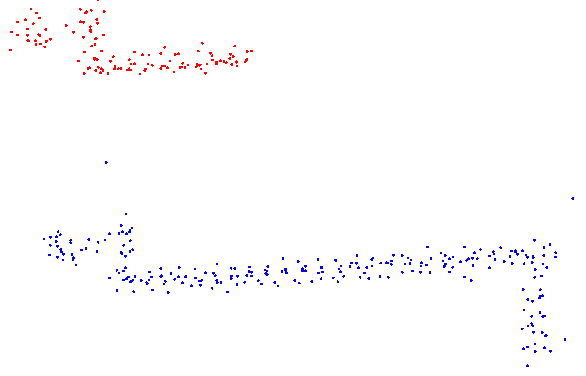}
\label{fig_reg_unreg}
}
\hspace{-2.0ex}
\subfloat[]
{
\includegraphics[width=0.15\linewidth,height=8ex,frame]{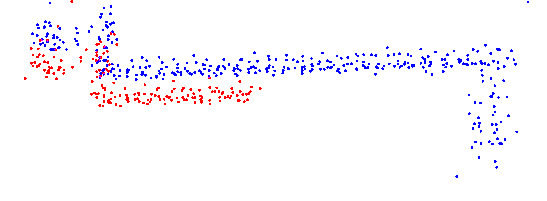}
\label{fig_reg_baseline}
}
\hspace{-2.0ex}
\subfloat[]
{
\includegraphics[width=0.15\linewidth,height=8ex,frame]{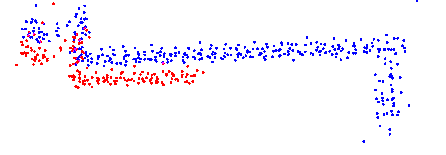}
\label{fig_reg_up_0}
}
\hspace{-2.0ex}
\subfloat[]
{
\includegraphics[width=0.15\linewidth,height=8ex,frame]{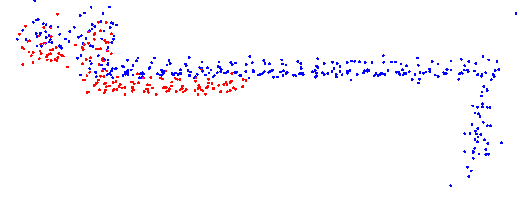}
\label{fig_reg_up_1}
}
\hspace{-2.0ex}
\subfloat[]
{
\includegraphics[width=0.15\linewidth,height=8ex,frame]{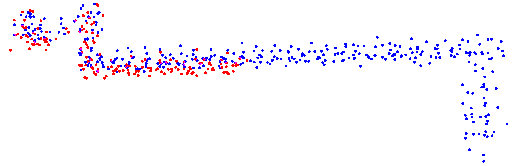}
\label{fig_reg_up_2}
}
\hspace{-2.0ex}
\subfloat[]
{
\includegraphics[width=0.15\linewidth,height=8ex,frame]{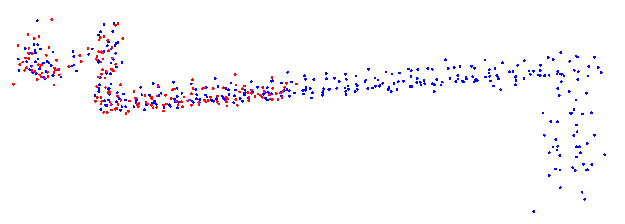}
\label{fig_reg_up_3}
}
\\
\subfloat[]
{
\includegraphics[width=0.45\linewidth,height=8ex]{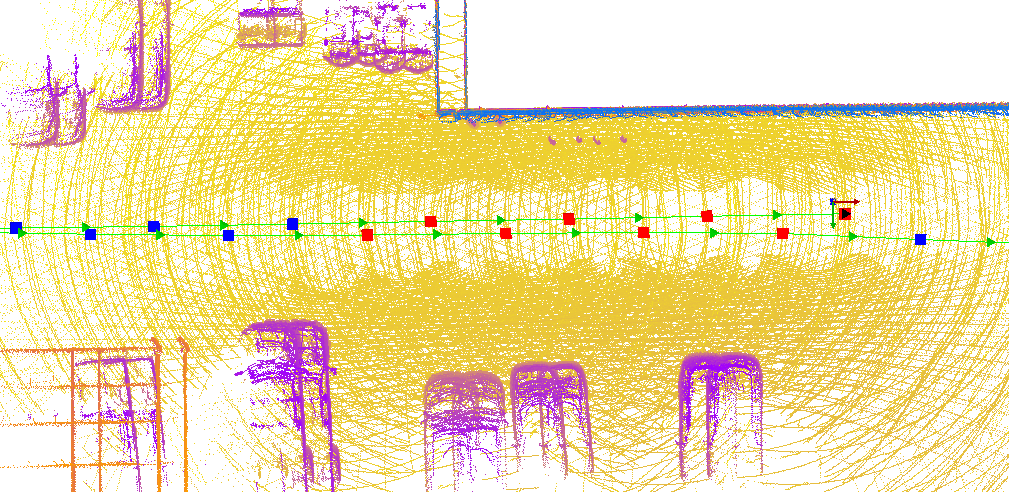}
\label{fig_without_pose_refinement}
}
\hfill
\subfloat[]
{
\includegraphics[width=0.45\linewidth,height=8ex]{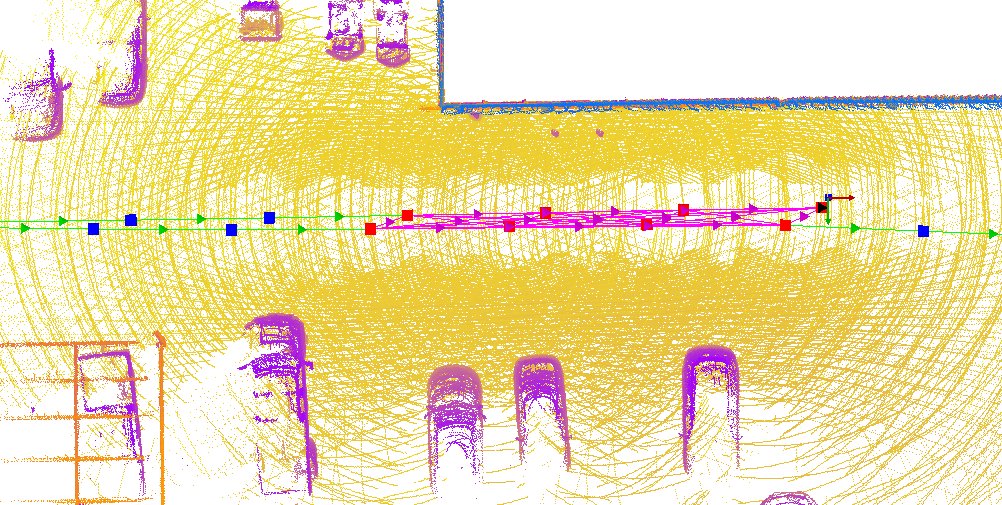}
\label{fig_with_pose_refinement}
}
\caption{ (a) raw scan-pair (red-blue), (b) aligned scans using baseline GICP, (c)-(f) \texttt{pose-refinement} for upsampling rates 0, 1, 2, and 3 respectively. Figure (g) an accumulated point cloud of an area by using raw poses (odometry) and (h) refined poses using \texttt{pose-refinement}.}
\label{fig_pose_refinement}
\end{figure}

Typically, a LIDAR can provide 3D world measurements in a radii of $\sim100m$ and these measurements are made w.r.t a vehicle/body frame ($V$). Hence for larger maps, the LIDAR scans need to be accumulated w.r.t. a fixed/local reference frame ($L$). However, the relative position of $V$ w.r.t. $L$ i.e. pose
, is governed by the GPS/INS data which itself is noisy and unreliable. Hence, scan accumulation by using the raw poses leads to very unpleasant 3D structures (car, building, wall) in the output map (Fig. \ref{fig_without_pose_refinement}).
\par
We handle this by performing \texttt{pose-refinement} where we aim to register pair of LIDAR scans so that the registered clouds can exhibit high degree of coherent surface \cite{harary2014context}. In order to achieve this, first we perform constrained upsampling of each scan i.e. upsample only the non-ground points. We perform this minor yet useful tweak due to the fact that the ground plane covers major area in a LIDAR scan and an upsampled ground plane may leave the registration algorithm trapped in local minima. In our case, we extract the ground plane by using $z$ thresholding as it is fast and works well at least locally. However, other approaches such as plane extraction using RANSAC \cite{fischler1987random} can also be exploited.  
\par
Although, upsampling a point cloud is non-trivial, we devise a fast and intuitive way to achieve this. First, we obtain image projection (pixel location) of LIDAR points using pin hole camera model and triangulate all the image projections using Delaunay triangulation. Later, for each triangle $t$ in pixel space, we retrieve a triangle $T$ in 3D space. Now, we insert a new vertex which is the image projection of the centroid of the triangle $T$. It is noteworthy that this operation essentially upsamples the point cloud due to insertion of new vertices into the Delaunay triangles and number of repetition of this step is equal to upsampling rate. We perform the insertion operation only for the triangles $T$ whose edges are below a threshold $\tau=0.3m$ as otherwise it may give rise to unwanted edges which physically are not present. The above discussed upsampling technique obtains a smooth surface in negligible time as compared to upsampling using \cite{mls} (Table \ref{tab_upsampling}). The smoothness in the point cloud is evident from the fact that we use averaging operation while inserting a new vertex into the Delaunay triangles.
 \begin{table}[!h]  
\centering        
\captionsetup{justification=centering}
 \caption{\\ \footnotesize Timing performance (CPU) of upsampling using proposed and \cite{mls}}
\scriptsize 
\begin{tabular}{| c | c | c | c | c | c |}
\hline
order of points & $10k$ & $30k$ & $90k$ & $180k$ & $450k$  \\ \cline{1-6}    
\cite{mls} & $2s$ & $15s$ & $160s$ & $1450s$ & $10000s$  \\ \cline{1-6}    
proposed & $30ms$ & $60ms$ & $120ms$ & $300ms$ & $900ms$  \\ \cline{1-6}    
 \end{tabular}
 \label{tab_upsampling}
 \end{table}
\par
Further, the upsampled scans are aligned (registered) using GICP \cite{gicp} with the raw poses as initial guess. 
From fig. \ref{fig_pose_refinement}, it can be seen that the constrained upsampling leads to a coherent registered surface which is rich in contextual 3D information such as 3D edges, corners, walls etc. We also verify this experimentally in sec. \ref{experiments}. 
\par
As we have discussed that the \texttt{pose-refinement} also uses GICP as underlying scan matching technique, we differentiate between pose estimation using standalone GICP and \texttt{pose-refinement} by referring the former as a ``baseline''. We maintain this keyword throughout the paper.

\subsection{Foveal Processing}
\label{subsec_foveal}
\begin{figure}[t]
\centering

\FPeval{\arrowlen}{4}
\FPeval{\largeangle}{100}

\FPeval{\camwidth}{2}
\FPeval{\camheight}{1}

\vspace{2ex}
\subfloat[Top view]
{
\begin{tikzpicture}

\FPeval{\imwidth}{10}
\FPeval{\imheight}{6}

\node (car) [rectangle,draw=white,minimum width= 10ex, minimum height= 10ex]{\includegraphics[width=\imwidth ex,height=\imheight ex]{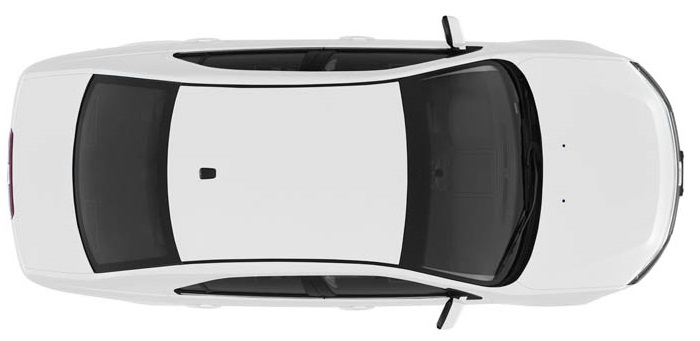}};


\FPeval{\camxshift}{0-1.0}

\foreach \i in {0,1,2,3,4}
{
\FPeval{\camxoffset}{clip(2.5*cos(2*pi*\i/5))}
\FPeval{\camyoffset}{clip(2.5*sin(2*pi*\i/5))}

\ifthenelse{\i=0}
{\FPeval{\smallangle}{10}}
{\FPeval{\smallangle}{80}}

\FPeval{\angleshift}{\i*360/5}
\FPeval{\minangle}{\angleshift-\largeangle/2}
\FPeval{\maxangle}{\angleshift+\largeangle/2}
\FPeval{\xoffset}{clip(\arrowlen*cos((2*pi*\i/5)-pi*\largeangle/360))}
\FPeval{\yoffset}{clip(\arrowlen*sin((2*pi*\i/5)-pi*\largeangle/360))}
\filldraw[fill=red!20!white, draw=red!90!black,opacity=0.7] (\camxshift ex+\camxoffset ex,\camyoffset ex) -- (\xoffset ex + \camxshift ex + \camxoffset ex, \yoffset ex+\camyoffset ex)  arc (\minangle:\maxangle:\arrowlen ex) -- ( \camxshift ex+\camxoffset ex,\camyoffset ex);
}

\foreach \i in {0,1,2,3,4}
{
\FPeval{\camxoffset}{clip(2.5*cos(2*pi*\i/5))}
\FPeval{\camyoffset}{clip(2.5*sin(2*pi*\i/5))}

\ifthenelse{\i=0}
{\FPeval{\smallangle}{50}}
{\FPeval{\smallangle}{10}}

\FPeval{\angleshift}{\i*360/5}
\FPeval{\minangle}{\angleshift-\smallangle/2}
\FPeval{\maxangle}{\angleshift+\smallangle/2}
\FPeval{\xoffset}{clip(\arrowlen*cos((2*pi*\i/5)-pi*\smallangle/360))}
\FPeval{\yoffset}{clip(\arrowlen*sin((2*pi*\i/5)-pi*\smallangle/360))}
\filldraw[fill=green!20!white, draw=green!90!black,opacity=0.7] (\camxshift ex+\camxoffset ex,\camyoffset ex) -- (\xoffset ex + \camxshift ex +\camxoffset ex,\yoffset ex+\camyoffset ex) arc (\minangle:\maxangle:\arrowlen ex) -- (\camxshift ex+\camxoffset ex,\camyoffset ex);
}

\foreach \i in {0,1,2,3,4}
{
\FPeval{\xoffset}{clip(1.5*cos(2*pi*\i/5))}
\FPeval{\yoffset}{clip(1.5*sin(2*pi*\i/5))}
\FPeval{\rotangle}{75*\i}
\node (cam_\i) [rectangle,minimum width= 4ex, minimum height=3ex, xshift=\camxshift ex+\xoffset ex,yshift=\yoffset ex]{\includegraphics[width=\camwidth ex,height=\camheight ex,angle=\rotangle]{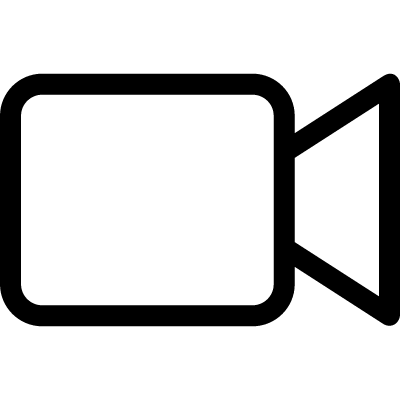}};
}
\FPeval{\blindradii}{10}
\FPeval{\whiteradii}{12}
\FPeval{\blindradiifar}{16}

\node (blindzone) [circle,draw=black,fill=white!40!black,opacity=0.2,minimum width= \blindradii ex, minimum height= \blindradii ex,xshift=\camxshift ex]{};

\node (whitezone) [circle,draw=cyan!100!white,line width=2.0ex, opacity=0.5, minimum width= \whiteradii ex, minimum height= \whiteradii ex,xshift= \camxshift ex]{};

\node (blindzonefar) [circle,draw=black,line width=2ex, color=white!40!black,opacity=0.2, minimum width= \blindradiifar ex, minimum height= \blindradiifar ex,xshift= \camxshift ex]{};

\node (restricted_fov) [rectangle, fill=green!20!white, xshift=-2ex +  \camxshift ex ,yshift=11ex]{\scriptsize Restricted FOV};
\node (restricted_fov_arrow) [rectangle, xshift=4ex +  \camxshift ex ,yshift=-1ex]{};
\draw [->,] (restricted_fov) -| (restricted_fov_arrow);

\end{tikzpicture}
\label{fig_cross_section_scanning_horz}
}
\hspace{5ex}
\subfloat[Side view]
{
\begin{tikzpicture}

\FPeval{\imwidth}{10}
\FPeval{\imheight}{5}

\node (car) [rectangle,draw=white,minimum width= 10ex, minimum height= 10ex]{\includegraphics[width=\imwidth ex,height=\imheight ex]{car_side.jpg}};


\FPeval{\camxshift}{0.0-0.5}
\FPeval{\camyshift}{2.0}

\foreach \i in {0,1}
{
\FPeval{\camxoffset}{clip(2*cos(2*pi*\i/2))}
\FPeval{\camyoffset}{clip(2*sin(2*pi*\i/2))}
\FPeval{\rotangle}{180*\i}
\node (cam_\i) [rectangle,minimum width= 4ex, minimum height=3ex, xshift= \camxoffset ex + \camxshift ex, yshift= \camyoffset ex + \camyshift ex]{\includegraphics[width=\camwidth ex,height=\camheight ex,angle=\rotangle]{camera.png}};
}

\FPeval{\smallangle}{10}

\foreach \i in {0,1}
{

\FPeval{\camxoffset}{clip(3.0*cos(2*pi*\i/2))}
\FPeval{\camyoffset}{clip(3.0*sin(2*pi*\i/2))}

\FPeval{\angleshift}{\i*360/2}
\FPeval{\minangle}{\angleshift-\largeangle/2}
\FPeval{\maxangle}{\angleshift+\largeangle/2}
\FPeval{\xoffset}{clip(\arrowlen*cos((2*pi*\i/2)-pi*\largeangle/360))}
\FPeval{\yoffset}{clip(\arrowlen*sin((2*pi*\i/2)-pi*\largeangle/360))}
\filldraw[fill=red!20!white, draw=red!90!black,opacity=0.7] (\camxoffset ex + \camxshift ex, \camyoffset ex + \camyshift ex) -- (\xoffset ex +\camxshift ex +\camxoffset ex,\yoffset ex +  \camyshift ex + \camyoffset ex) arc (\minangle:\maxangle:\arrowlen ex) -- (\camxoffset ex + \camxshift ex, \camyoffset ex + \camyshift ex);
}

\foreach \i in {0,1}
{

\FPeval{\camxoffset}{clip(3.0*cos(2*pi*\i/2))}
\FPeval{\camyoffset}{clip(3.0*sin(2*pi*\i/2))}

\ifthenelse{\i=0}
{
\FPeval{\smallangle}{10}
\FPeval{\angleshift}{345}
}
{
\FPeval{\smallangle}{50}
\FPeval{\angleshift}{180}
}

\FPeval{\minangle}{\angleshift-\smallangle/2}
\FPeval{\maxangle}{\angleshift+\smallangle/2}
\FPeval{\xoffset}{clip(\arrowlen*cos((pi*\angleshift/180)-(pi*\smallangle/360)))}
\FPeval{\yoffset}{clip(\arrowlen*sin((pi*\angleshift/180)-(pi*\smallangle/360)))}
\filldraw[fill=green!20!white, draw=green!90!black,opacity=0.7] (\camxoffset ex + \camxshift ex, \camyoffset ex + \camyshift ex) -- (\xoffset ex +\camxshift ex +\camxoffset ex,\yoffset ex +  \camyshift ex + \camyoffset ex) arc (\minangle:\maxangle:\arrowlen ex) -- (\camxoffset ex + \camxshift ex, \camyoffset ex + \camyshift ex);
}

\FPeval{\zoneminangle}{210}
\FPeval{\zonemaxangle}{0-30}

\FPeval{\blindradii}{6}
\FPeval{\whiteradii}{7}
\FPeval{\blindradiifar}{9}

\FPeval{\zonexoffset}{clip(\blindradii *cos(pi*\zoneminangle/180))}
\FPeval{\zoneyoffset}{clip(\blindradii *sin(pi*\zoneminangle/180))}

\filldraw [draw=black,fill=white!40!black, opacity=0.2
,preaction={decorate,decoration={text along path,text={|\tiny|near blind zone},text align={align=center}, raise=-1.0ex}}
]
(\camxshift ex,\camyshift ex) -- (\zonexoffset ex + \camxshift ex ,\zoneyoffset ex +\camyshift ex ) arc (\zoneminangle:\zonemaxangle:\blindradii ex) -- (\camxshift ex,\camyshift ex);

\FPeval{\annularradii}{\whiteradii-\blindradii}

\FPeval{\outerzonexoffset}{clip((\blindradii + \annularradii)*cos(pi*\zoneminangle/180))}
\FPeval{\outerzoneyoffset}{clip((\blindradii + \annularradii)*sin(pi*\zoneminangle/180))}

\draw [draw=cyan!100!white, opacity=0.5, line width=2 ex
,preaction={decorate,decoration={text along path,text={|\tiny|white zone},text align={align=center},raise=-0.3ex}}
]
(\outerzonexoffset ex + \camxshift ex ,\outerzoneyoffset ex +\camyshift ex ) arc (\zoneminangle:\zonemaxangle:\whiteradii ex);

\FPeval{\annularradii}{\blindradiifar -\whiteradii}

\FPeval{\outerzonexoffset}{clip((\whiteradii + \annularradii)*cos(pi*\zoneminangle/180))}
\FPeval{\outerzoneyoffset}{clip((\whiteradii + \annularradii)*sin(pi*\zoneminangle/180))}

\draw [draw=white!40!black, opacity=0.2, line width=2 ex
,preaction={decorate,decoration={text along path,text={|\tiny|far blind zone},text align={align=center},raise=-0.3ex}}
](\outerzonexoffset ex + \camxshift ex ,\outerzoneyoffset ex +\camyshift ex ) arc (\zoneminangle:\zonemaxangle:\blindradiifar ex);


\node (full_fov) [rectangle, fill=red!20!white,  xshift=0ex +  \camxshift ex ,yshift=15ex]{\scriptsize Actual FOV};

\node (full_fov_arrow) [rectangle, xshift=5ex +  \camxshift ex ,yshift=2ex]{};

\draw [->] (full_fov) -| (full_fov_arrow);

\end{tikzpicture}
\label{fig_cross_section_scanning_vert}
}
\caption{This figure shows the restricted field of view (FOV) of cameras, white zone, near and far blind zones used for texture mapping.}
\label{fig_cross_section_scanning}
\end{figure}

\begin{figure}[t]
\centering
\vspace{-2ex}
\subfloat[Frame $210$]
{
\includegraphics[width=0.30\linewidth,height=13ex]{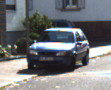}
}
\subfloat[Frame $220$]
{
\includegraphics[width=0.30\linewidth,height=13ex]{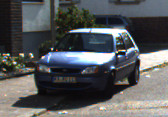}
}
\subfloat[Frame $228$]
{
\includegraphics[width=0.30\linewidth,height=13ex]{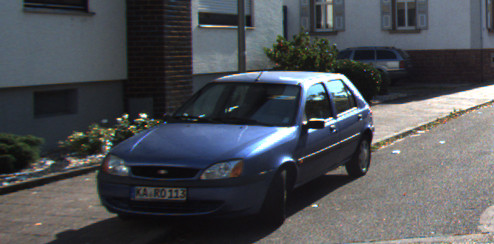}
}
\caption{Notice how the sunlight affects color texture of the car as the surveying vehicle navigates from figure (a) to (c). Frames taken from KITTI sequence $0095$.}
\label{fig_shade_change}
\end{figure}

Through experimental observation, we have noticed that the farther points does not receive a fine grained color texture due to sparsity, noise in LIDAR scans and large angle ($\geq\sim70^{o}$) between view point and normal at the point. Moreover, due to external factors e.g. sunlight, reflective surface, the color texture of the objects in the images changes drastically as the vehicle navigate through the environment (Fig. \ref{fig_shade_change}). In such cases, approaches such as \cite{romanoni2017mesh} doesn't preserve texture due to weighted averaging (kindly refer to \cite{romanoni2017mesh} for further details) in their texturing process. Hence, inspired from human fovea \cite{DEVOE1968135}, where major processing happens in a relatively smaller region of visual field, we define similar regions by restricting field of view (FOV) of the camera to a thin horizontal and a thin vertical slice (2D \texttt{foveal~region}). Moreover, we also define a spherical near blind, white and far blind zone (3D \texttt{foveal~region}) around the car (Fig. \ref{fig_cross_section_scanning}). The texture mapping is performed only in the white zone whereas it is avoided in the blind zones because the near blind zone encloses vehicle itself and the far blind zone have noisy depth measurements. The restricted FOV, various blind and white zones together improves the quality of texture mapping drastically and simultaneously improves algorithmic speed due to reduced number of points, which are to be processed.
\par
Further, as soon as an aligned scan becomes available, its foveal region is extracted (Fig. \ref{fig_foveal_extraction}) which is upsampled including the ground plane and is operated by the \texttt{ray-filtering} discussed below. 

\subsection{Ray-filtering and Texture Mapping}
\label{subsec_ray_filtering}
%
%
The \texttt{ray-filtering} operation plays an important role in obtaining a fine grained texture in the map. 
This operation discards the points $P_{S}$ in the scan $S$ which if added to the accumulated cloud (from previous scans), gets $occluded$ or $occludes$ any existing point in the accumulated cloud. In general, such situations arise due to inconsistencies in the pose and must be handled so as to maintain coherent 3D surface in the output map.
%
\par
\begin{figure}[t]
\centering
\captionsetup{justification=centering}
\vspace{1.5ex}
\subfloat[Case of occluded]
{
\tdplotsetmaincoords{0}{0}     

\begin{tikzpicture}[tdplot_main_coords,scale=1]
 \FPeval{\axislength}{5}
 \FPeval{\hypotaneous}{\axislength/1.414}

 \draw [->,very thick, color=blue] (0,0) -- (\axislength ex,0ex) node [pos=1.1]{$z$};
 \draw [->,very thick, color=green] (0,0) -- (0ex,\axislength ex)node [pos=1.1]{$y$};
 \draw [->,very thick, color=red] (0,0) -- (-\hypotaneous ex,-\hypotaneous ex)node [pos=1.1]{$x$};

 \node (o) [rectangle,minimum width= 1ex, minimum height= 1ex,xshift=-2ex]{C};
 \node (i) [rectangle,minimum width= 1ex, minimum height= 1ex,xshift=8ex,yshift=8ex]{\tiny I};
 \node (w) [rectangle,minimum width= 1ex, minimum height= 1ex,xshift=6ex,yshift=5.5ex]{\tiny W};




\FPeval{\boxwidth}{4}
\FPeval{\boxheight}{1}
\FPeval{\boxhypo}{\boxwidth*\boxwidth+\boxheight*\boxheight}
\FProot{\boxhypo}{\boxhypo}{2}

\FPeval{\bxshift}{3}
\FPeval{\byshift}{5}

\filldraw [draw=black, fill=black,opacity=0.2]
(-\boxhypo ex + \bxshift ex, -\boxhypo ex - \byshift ex) -- 
(\boxhypo ex  + \bxshift ex, \boxhypo ex - \byshift ex) -- 
(\boxhypo ex  + \bxshift ex, \boxhypo ex + \byshift ex ) --
(-\boxhypo ex + \bxshift ex, -\boxhypo ex +\byshift ex) --
(-\boxhypo ex + \bxshift ex, -\boxhypo ex - \byshift ex); 

\FPeval{\boxwidth}{2.5}
\FPeval{\boxheight}{1}
\FPeval{\boxhypo}{\boxwidth*\boxwidth+\boxheight*\boxheight}
\FProot{\boxhypo}{\boxhypo}{2}

\FPeval{\bxshift}{0.6*5}
\FPeval{\byshift}{2}

\filldraw [draw=green, fill=black,opacity=0.2]
(-\boxhypo ex + \bxshift ex, -\boxhypo ex - \byshift ex) -- 
(\boxhypo ex  + \bxshift ex, \boxhypo ex - \byshift ex) -- 
(\boxhypo ex  + \bxshift ex, \boxhypo ex + \byshift ex ) --
(-\boxhypo ex + \bxshift ex, -\boxhypo ex +\byshift ex) --
(-\boxhypo ex + \bxshift ex, -\boxhypo ex - \byshift ex);

\foreach \i in {-3,...,3}
\foreach \j in {-1,...,1}
{
\node (p_\i\j) [draw=cyan, fill=cyan!40!white,circle,minimum width=1ex,minimum height=1ex,scale = 0.4,xshift=\i*3ex+ 30ex,yshift=\j*2.5ex+\i*3ex]{};
}

\foreach \i in {-1,...,1}
\foreach \j in {0}
{
\node (q_\i\j) [draw=red, fill=red!40!white,circle,minimum width=1ex,minimum height=1ex,scale = 0.4,xshift=\i*3ex+\j*4ex+ 40ex,yshift=\i*3ex]{};
}

\foreach \i in {-3,...,3}
\foreach \j in {-1,...,1}
{
\FPeval{\pnxshift}{(0.9)*(\i*4)+ 15}
\FPeval{\pnyshift}{(0.9)*(\j*3 + \i*4)}

\node (pn_\i\j) [draw=cyan, fill=cyan!40!white,circle,minimum width=1ex,minimum height=1ex,scale = 0.2,xshift=\pnxshift ex,yshift=\pnyshift ex]{};
}

\foreach \i in {-1,...,1}
\foreach \j in {0}
{
\FPeval{\pnxshift}{(0.9)*(\i*4)+ 15}
\FPeval{\pnyshift}{(0.9)*(\j*3 + \i*4)}

\node (qn_\i\j) [draw=red, fill=red!40!white,circle,minimum width=1ex,minimum height=1ex,scale = 0.2,xshift=\pnxshift ex,yshift=\pnyshift ex]{};
}

\foreach \i in {-3,...,3}
\foreach \j in {-1}
{
\draw [->,color=cyan!40!white] (p_\i\j) -- (pn_\i\j);
}

\foreach \i in {-1,...,1}
\foreach \j in {0}
{
\draw [->,color=red!40!white] (q_\i\j) -- (qn_\i\j);
}
\end{tikzpicture}
}
\hspace{-2.5ex}
\subfloat[Case of occluding]
{
\tdplotsetmaincoords{0}{0}     

\begin{tikzpicture}[tdplot_main_coords,scale=1]
 \FPeval{\axislength}{5}
 \FPeval{\hypotaneous}{\axislength/1.414}

 \draw [->,very thick, color=blue] (0,0) -- (\axislength ex,0ex) node [pos=1.1]{$z$};
 \draw [->,very thick, color=green] (0,0) -- (0ex,\axislength ex)node [pos=1.1]{$y$};
 \draw [->,very thick, color=red] (0,0) -- (-\hypotaneous ex,-\hypotaneous ex)node [pos=1.1]{$x$};

 \node (o) [rectangle,minimum width= 1ex, minimum height= 1ex,xshift=-2ex]{C};
 \node (i) [rectangle,minimum width= 1ex, minimum height= 1ex,xshift=8ex,yshift=8ex]{\tiny I};
 \node (w) [rectangle,minimum width= 1ex, minimum height= 1ex,xshift=6ex,yshift=5.5ex]{\tiny W};




\FPeval{\boxwidth}{4}
\FPeval{\boxheight}{1}
\FPeval{\boxhypo}{\boxwidth*\boxwidth+\boxheight*\boxheight}
\FProot{\boxhypo}{\boxhypo}{2}

\FPeval{\bxshift}{3}
\FPeval{\byshift}{5}

\filldraw [draw=black, fill=black,opacity=0.2]
(-\boxhypo ex + \bxshift ex, -\boxhypo ex - \byshift ex) -- 
(\boxhypo ex  + \bxshift ex, \boxhypo ex - \byshift ex) -- 
(\boxhypo ex  + \bxshift ex, \boxhypo ex + \byshift ex ) --
(-\boxhypo ex + \bxshift ex, -\boxhypo ex +\byshift ex) --
(-\boxhypo ex + \bxshift ex, -\boxhypo ex - \byshift ex); 

\FPeval{\boxwidth}{2.5}
\FPeval{\boxheight}{1}
\FPeval{\boxhypo}{\boxwidth*\boxwidth+\boxheight*\boxheight}
\FProot{\boxhypo}{\boxhypo}{2}

\FPeval{\bxshift}{0.6*5}
\FPeval{\byshift}{2}

\filldraw [draw=green, fill=black,opacity=0.2]
(-\boxhypo ex + \bxshift ex, -\boxhypo ex - \byshift ex) -- 
(\boxhypo ex  + \bxshift ex, \boxhypo ex - \byshift ex) -- 
(\boxhypo ex  + \bxshift ex, \boxhypo ex + \byshift ex ) --
(-\boxhypo ex + \bxshift ex, -\boxhypo ex +\byshift ex) --
(-\boxhypo ex + \bxshift ex, -\boxhypo ex - \byshift ex);

\foreach \i in {-3,...,3}
\foreach \j in {-1,...,1}
{
\node (p_\i\j) [draw=cyan, fill=cyan!40!white,circle,minimum width=1ex,minimum height=1ex,scale = 0.4,xshift=\i*3ex+ 40ex,yshift=\j*2.5ex+\i*3ex]{};
}

\foreach \i in {-1,...,1}
\foreach \j in {0}
{
\node (q_\i\j) [draw=red, fill=red!40!white,circle,minimum width=1ex,minimum height=1ex,scale = 0.4,xshift=\i*3ex+\j*4ex+ 30ex,yshift=\i*3ex]{};
}

\foreach \i in {-3,...,3}
\foreach \j in {-1,...,1}
{
\FPeval{\pnxshift}{(0.9)*(\i*4)+ 15}
\FPeval{\pnyshift}{(0.9)*(\j*3 + \i*4)}

\node (pn_\i\j) [draw=cyan, fill=cyan!40!white,circle,minimum width=1ex,minimum height=1ex,scale = 0.2,xshift=\pnxshift ex,yshift=\pnyshift ex]{};
}

\foreach \i in {-1,...,1}
\foreach \j in {0}
{
\FPeval{\pnxshift}{(0.9)*(\i*4)+ 15}
\FPeval{\pnyshift}{(0.9)*(\j*3 + \i*4)}

\node (qn_\i\j) [draw=red, fill=red!40!white,circle,minimum width=1ex,minimum height=1ex,scale = 0.2,xshift=\pnxshift ex,yshift=\pnyshift ex]{};
}

\foreach \i in {-3,...,3}
\foreach \j in {-1}
{
\draw [->,color=cyan!40!white] (p_\i\j) -- (pn_\i\j);
}

\foreach \i in {-1,...,1}
\foreach \j in {0}
{
\draw [->,color=red!40!white] (q_\i\j) -- (qn_\i\j);
}

\end{tikzpicture}
}
\caption{The figure above describes different ways, the points can be occluded or occluding}
\label{fig_ray_filtering}
\end{figure}
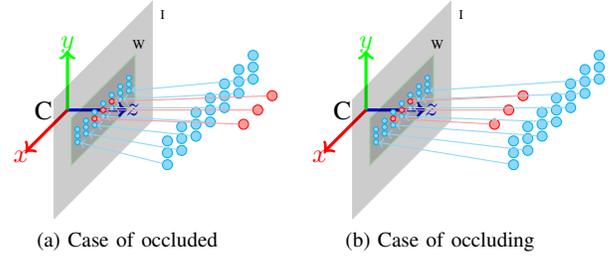

%
To achieve this, first we extract all the points $P_{A}^{f}$ that lie in the \texttt{foveal~region} of current $pose$. Next, a virtual ray from each point $P$ in $\{P_{S},~P_{A}^{f}\}$ is emanated towards camera origin (Fig. \ref{fig_ray_filtering}) and all of the such rays are projected into the camera using the pin-hole camera model\footnote{$p=K[R|t]P$, where $K$, $R~\vert~t$ are intrinsic and extrinsic camera matrix}. However, collinearity of rays or high density around a point sometimes result in a common pixel projection ($p$) for multiple rays. To handle this case, we assign $p$ to a ray having shortest ray-length $d$ among all such rays. If $d$ corresponds to a $P_{S}$, it is discarded as this happens to be the case of $occluding$.
%
%
%
%

\par
Further, to test the case of $occluded$, we take a window (W) of size $[M \times M]$ centered at the image projection $p$ of each $P_{S}$ (Fig. \ref{fig_ray_filtering}). Now, we compute mean ($\mu$) and standard deviation ($\sigma$) (Eq. \ref{eq:mean_std_dev}) of all the ray-lengths $d_{i}$s inside the window and perform a statistical outlier rejection test (Eq. \ref{eq:statistical_outlier}) on them in order to obtain an outlier score. If the outlier score is less than an outlier rate $c$, the point $P_t$ is marked as visible otherwise it is considered as occluded. In general, lesser the value of $c$, higher will be the rejection rate i.e. more number of points will be marked as occluded. Hence, frequency of inliers in a window can be controlled by varying the value of $c$. 
\begin{equation}
\begin{aligned}
\mu &= \frac{1}{N} \sum_{i=1}^{N} d_{i}  \\
\sigma^{2} &= \frac{1}{N} \sum_{i=1}^{N} \left \Vert d_{i}-\mu) \right \Vert^2 
\end{aligned}
\label{eq:mean_std_dev}
\end{equation}
\begin{equation}
P_{S}~\text{is} =
\begin{cases}
Visible,  & \text{if } \frac{|d_{i}-\mu|}{\sigma} \leq ~c \\
Occluded, & \text{Otherwise}
\end{cases}
\label{eq:statistical_outlier}
\end{equation}
Where, $N \leqslant M \times M $ is total number of rays projected in the window. 
\par
\begin{figure}[t]
\centering
\subfloat[Frame $0$]
{
\includegraphics[width=0.30\linewidth,height=8ex]{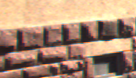}
}
\hfill
\subfloat[Frame $0$]
{
\includegraphics[width=0.30\linewidth,height=8ex]{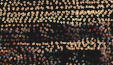}
}
\hfill
\subfloat[Frame $0$]
{
\includegraphics[width=0.30\linewidth,height=8ex]{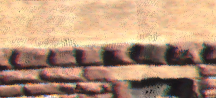}
}
\vspace{-1.5ex}
\subfloat[Frame $20$]
{
\includegraphics[width=0.30\linewidth,height=8ex]{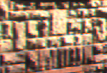}
}
\hfill
\subfloat[Frame $20$]
{
\includegraphics[width=0.30\linewidth,height=8ex]{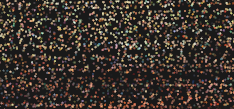}
}
\hfill
\subfloat[Frame $20$]
{
\includegraphics[width=0.30\linewidth,height=8ex]{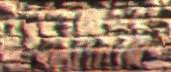}
}
\vspace{-1.5ex}

\subfloat[Frame $20-40$]
{
\includegraphics[width=0.47\linewidth,height=15ex]{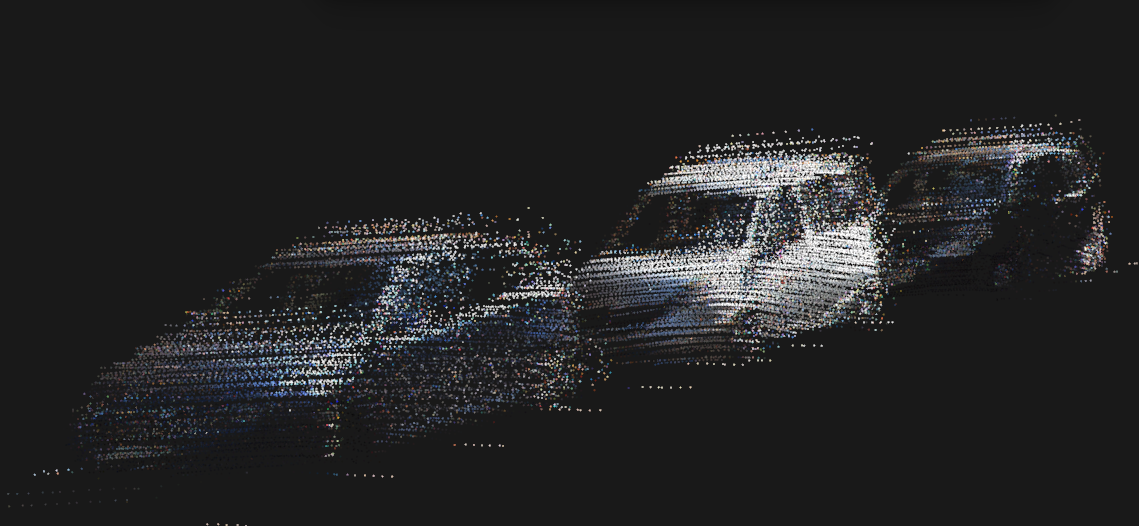}
}
\hfill
\subfloat[Frame $20-40$]
{
\includegraphics[width=0.47\linewidth,height=15ex]{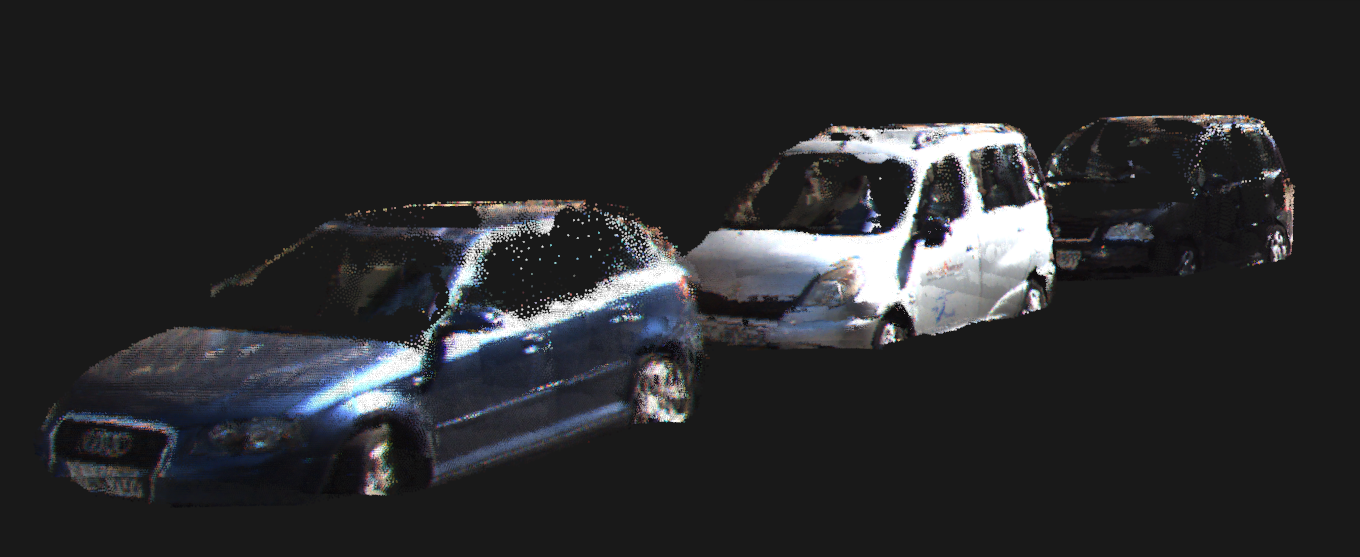}
}
\caption{Zoom in for better insight. Figure (a), (d) ground truth patch images. Corresponding patches of textured point cloud using (b), (e) baseline texturing and (c), (f) proposed texturing. (g) textured point cloud using baseline, and (h) using proposed texturing. Frames taken from KITTI sequence $0095$.}
\label{fig_texturing}
\end{figure}

Now, we texturize the filtered points from $S$ by projecting them into the image and later add them to the accumulated cloud. Unlike \cite{romanoni2017mesh,callieri2008masked} in which color texturing is a weighted averaging operation (low pass filtering) and doesn't preserve sharp edges, the visibility test using \texttt{ray-filtering} inherently preserves the image texture information resulting in fine grained texture detail in the textured map (Fig. \ref{fig_texturing}).
\par
If a $P_{S}$ passes both the tests, it represents the case of space carving. The \texttt{ray-filtering} performs it in an entirely different manner as no triangulation or any ray-intersection test is performed in contrast to \cite{litvinov2013incremental,litvinov2014incremental}, which triangulate the points and classify the 3D space as matter or void by ray-intersection.

%
%
%
%
%



%

\subsection{Error Metrics}
\label{subsec_error_metric}
Authors in the previous works \cite{romanoni2015incremental,romanoni2017mesh} have reported the quantitative $map~error$ ($ME$, Eq. \ref{eq_map_error}), 
which is mean ($\mu_{ME}$) and standard deviation ($\sigma_{ME}$) of euclidean distances between the points and their ground truth. 
This metric provides a measure for the quality of the 3D surface of the map but it does not convey anything about the quality of texture. Therefore, in this work we propose a quantitative error metric for texture quality of the map. We propose two error metrics (\textit{i}) $texture~error$ ($TE$, Eq. \ref{eq_texture_error}) and (\textit{ii}) $mean~texture~mapping~error$ ($MTME$, Eq. \ref{eq_mtme}). The former is the mean ($\mu_{ME}$) and standard deviation ($\sigma_{ME}$) of euclidean distances between the intensity/RGB of points and their ground truth. While the latter is an averaged sum of product of euclidean distances between 3D location and intensity/color of the points and their ground truth. For texture mapping purpose, $TE,~MTME$ are computed over all the points in the original LIDAR scans. 
Both $TE$ and $MTME$ provides quantifiable measures to assess the texture and overall quality of textured maps. We have experimentally demonstrated the usefulness of both $TE$ and $MTME$ in the section \ref{experiments} below.
\begin{equation}
\begin{aligned}
\mu_{ME} &= \frac{1}{N}~\sum_{i=1}^{n}~ \sum_{j=1}^{k_{i}}~ \left \Vert P_{j}^{L} - P_{n}^{M} \right \Vert \\
\sigma_{ME}^2 &= \frac{1}{N}~\sum_{i=1}^{n}~ \sum_{j=1}^{k_{i}}~ \left \Vert P_{j}^{L} - \mu_{ME} \right \Vert^{2}
\end{aligned}
\label{eq_map_error}
\end{equation}
\begin{equation}
\begin{aligned}
\mu_{TE} &= \frac{1}{N}~\sum_{i=1}^{n}~ \sum_{j=1}^{k_{i}}~ \left \Vert C_{j}^{L} - C_{n}^{M} \right \Vert \\
\sigma_{TE}^{2} &= \frac{1}{N}~\sum_{i=1}^{n}~ \sum_{j=1}^{k_{i}}~ \left \Vert C_{j}^{L} - \mu_{TE} \right \Vert^{2}  
\end{aligned}
\label{eq_texture_error}
\end{equation}
\begin{equation}
MTME = \frac{1}{N}~\sum_{i=1}^{n}~ \sum_{j=1}^{k_{i}}~\left \Vert P_{j}^{L} - P_{n}^{M}\right \Vert . \left \Vert C_{j}^{L} - C_{n}^{M}\right \Vert  
\label{eq_mtme}
\end{equation}
where 
\begin{description}
\item [$k_{i}$] total number of points in $i^{th}$ LIDAR scan  
\item [$P_{j}^{L}$] $j^{th}$ point in LIDAR scan
\item [$P_{n}^{M}$] nearest neighbor of $P_{j}^{L}$ in the map
\item [$C_{j}^{L}$] intensity $I$ or color $[R,G,B]$ of $P_{j}^{L}$
\item [$C_{n}^{M}$] intensity $I$ or color $[R,G,B]$ of $P_{n}^{M}$
\item [$\mu_{ME}$] mean of the map error
\item [$\sigma_{ME}$] standard deviation of the map error
\item [$\mu_{TE}$] mean of the texture error
\item [$\sigma_{TE}$] standard deviation of the texture error
\item [$n$] Total number of scans
\item [$N = \sum_{i=1}^{n}k_{i}$]
\end{description}
 %


\section{Experiments and Results}
\label{experiments}
We validate the performance of the proposed framework against the publicly available KITTI dataset \cite{geiger2013vision}. Our focus in this paper has remained on generating high resolution textured maps. Hence, from the dataset we only choose the sequences having minimal number of moving objects, in particular, sequence $0095$ ($268$ frames) and $0001$ ($108$ frames). Each sequence has time synchronized frames, each of which contains LIDAR scans captured by velodyne HDL-64E, RGB image of size 1242x375 captured by Point Grey Flea 2 and navigation data measurements by OXTS RT 3003 IMU/GPS. All the experiments were performed on a computing platform having 2 x Intel Xeon-2693 E5 CPUs, 256 GB RAM, 8 x NVIDIA Geforce GTX 1080Ti GPUs. 
%
%
\subsection{Scan Matching Performance and Timing Analysis}
\par
We estimate vehicle pose using \texttt{pose-refinement} and consider each of generalized-ICP (GEN-ICP) \cite{gicp}, standard-ICP (STD-ICP), and point-to-plane-ICP (P2P-ICP) as underlying scan matching technique. Due to unavailability of ground truth poses, we report point-to-point error between aligned scans using \texttt{pose-refinement} for upsampling rates 0xUp\footnote{camera visibility constraint but no upsampling}, 1xUp, 2xUp, 3xUp. We also report timing performance (CPU) averaged over the number of frames in the respective sequence. We compare our results with those achieved by the standalone baseline GEN-ICP, STD-ICP, and P2P-ICP, the navigation data serves as initial guess for all the experiments 

\par
\begin{figure}
\centering
\subfloat[]
{
\includegraphics[width=0.22\linewidth,height=8ex,frame]{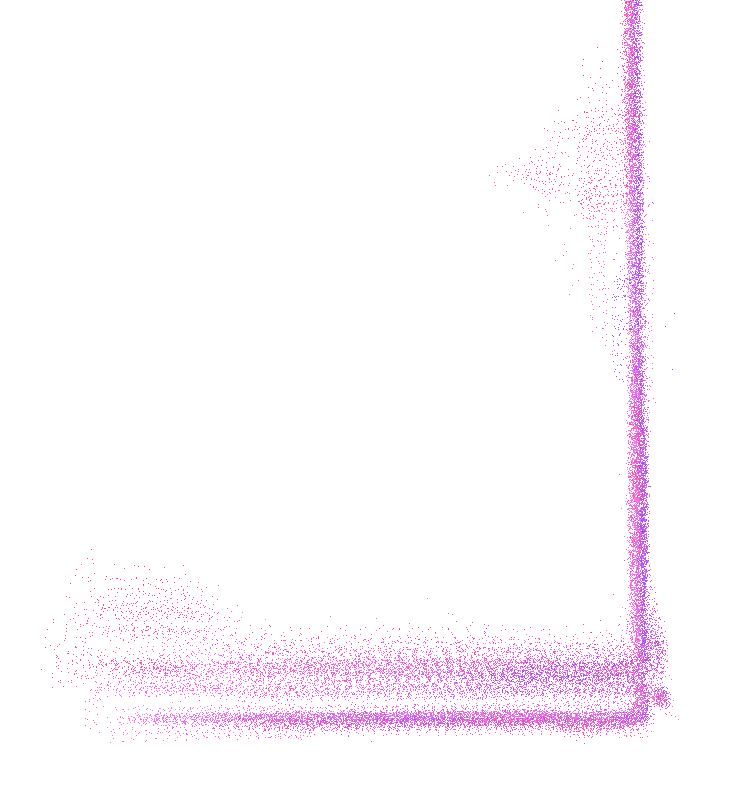}
}
\hspace{-2.0ex}
\subfloat[]
{
\includegraphics[width=0.22\linewidth,height=8ex,frame]{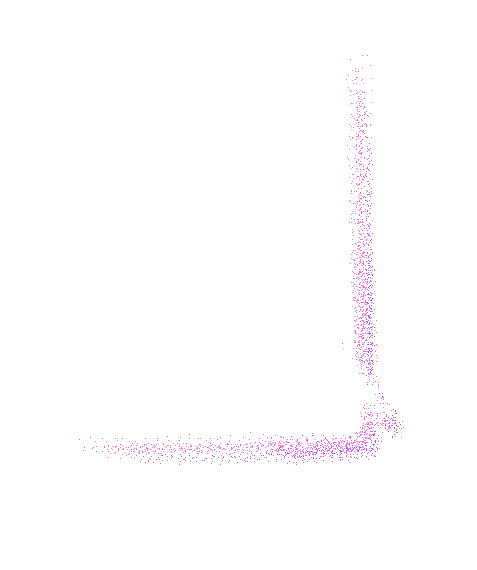}
}
\hfill
\subfloat[]
{
\includegraphics[width=0.22\linewidth,height=8ex,frame]{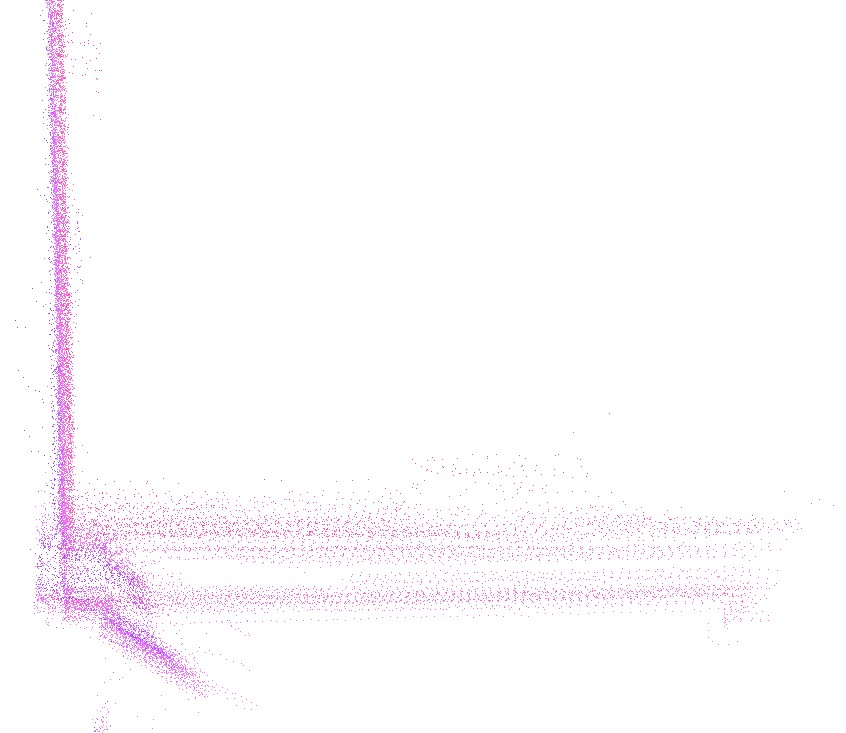}
}
\hspace{-2.0ex}
\subfloat[]
{
\includegraphics[width=0.22\linewidth,height=8ex,frame]{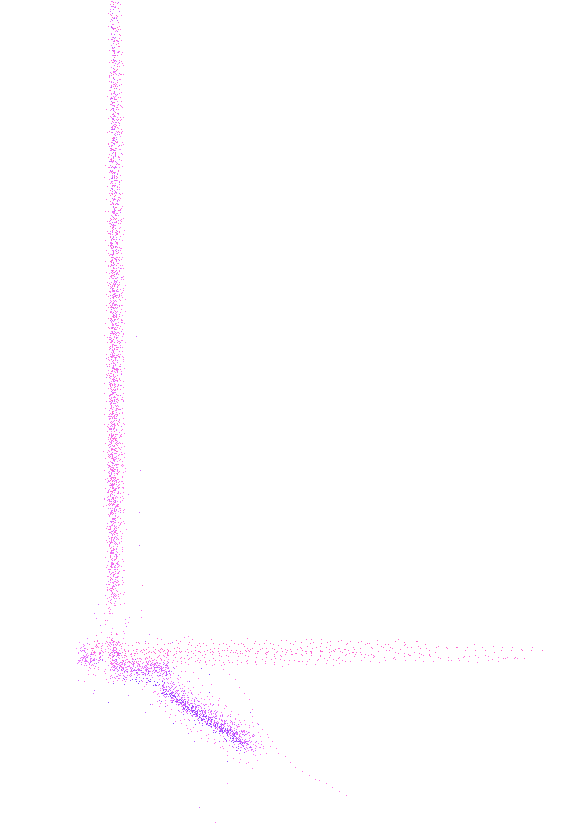}
}
\vspace{-0.0ex}
\subfloat[]
{
\includegraphics[width=0.22\linewidth,height=8ex,frame]{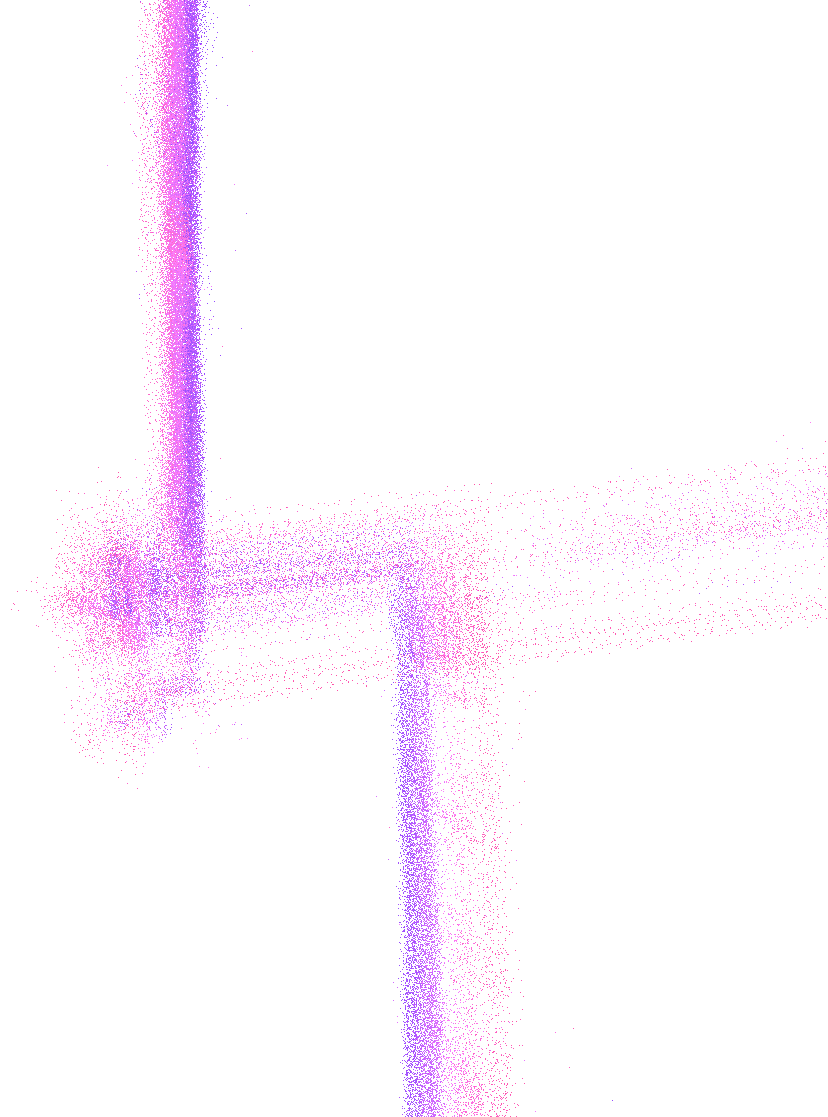}
}
\hspace{-2.0ex}
\subfloat[]
{
\includegraphics[width=0.22\linewidth,height=8ex,frame]{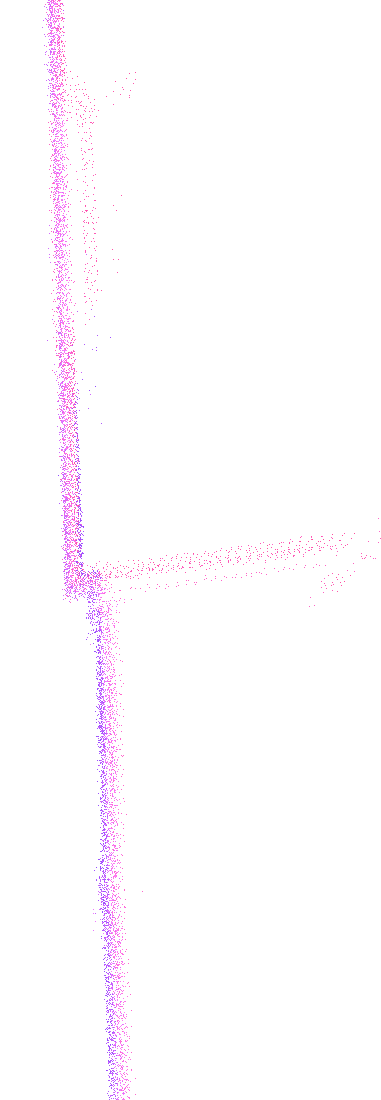}
}
\hfill
\subfloat[]
{
\includegraphics[width=0.22\linewidth,height=8ex,frame]{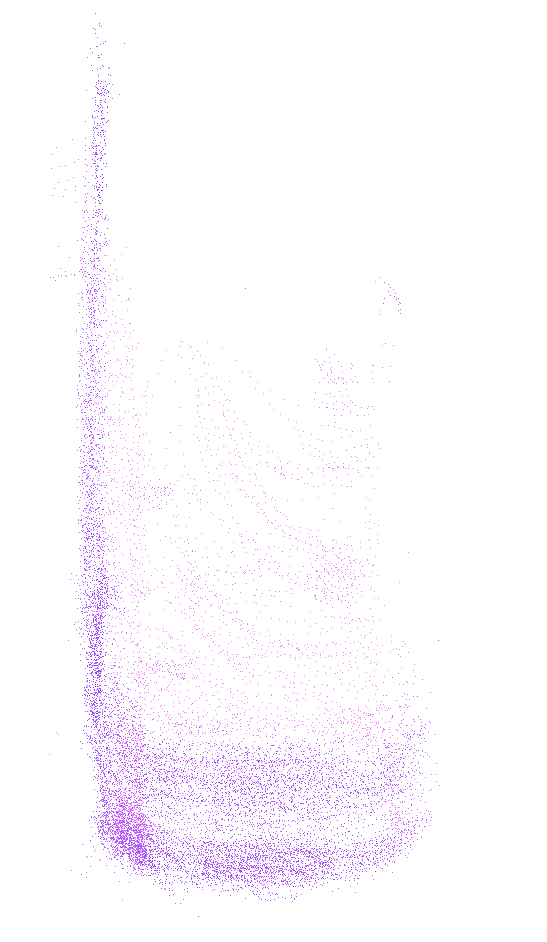}
}
\hspace{-2.0ex}
\subfloat[]
{
\includegraphics[width=0.22\linewidth,height=8ex,frame]{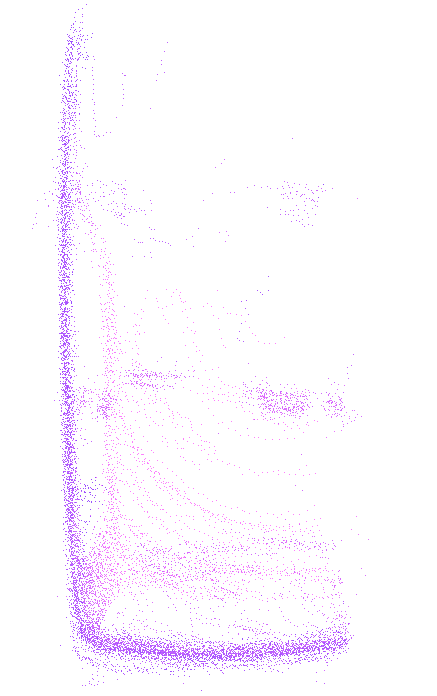}
}
\caption{(a),(c),(e),(g) scan aligned using baseline GICP, and (b),(d),(f),(h) \texttt{pose-refinement} with GICP + 3xUp }
\label{fig_reg_quality}
\end{figure}
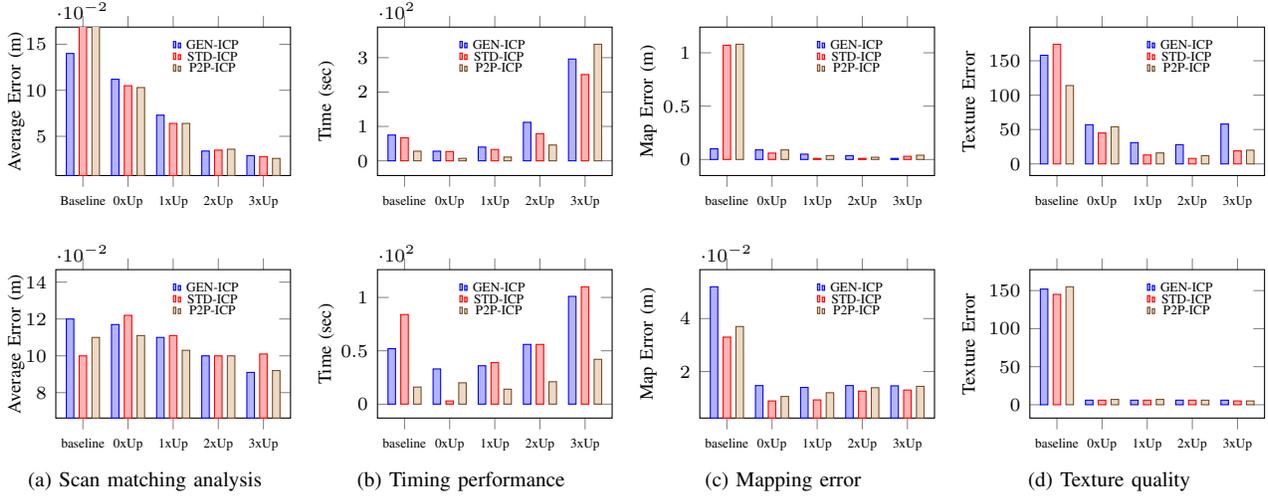
\begin{figure*}
\centering
\captionsetup{justification=centering}
\subfloat
{
\begin{tikzpicture}
\begin{axis}[
  ybar,
    x=0.6cm,
    bar width=0.1cm,
    enlargelimits=0.15,
    ymax = 0.15,
    width=0.24\textwidth,
     height=.20\textwidth,
    legend style={at={(0.65,0.99)}, anchor=north,font=\tiny, draw=none, row sep=-1ex},
    legend image post style={scale=0.3},
     y label style={at={(axis description cs:.30,.5)},rotate=0,anchor=south},
    label style={font=\scriptsize},
    x tick label style={font=\tiny},
    y tick label style={font=\scriptsize},
    ylabel={Average Error (m)},
    symbolic x coords={Baseline, 0xUp, 1xUp, 2xUp, 3xUp },
    xtick=data,
    scaled y ticks=base 10:2
    ]
\addplot coordinates {(Baseline,.14) (0xUp,.112) (1xUp,.073) (2xUp,.034) (3xUp,.029) };
\addplot coordinates {(Baseline,3.15) (0xUp,.105) (1xUp,.064) (2xUp,.035) (3xUp,.028) };
\addplot coordinates {(Baseline,4.14) (0xUp,.103) (1xUp,.064) (2xUp,.036) (3xUp,.026) };

\legend{GEN-ICP, STD-ICP, P2P-ICP}
\end{axis}
\end{tikzpicture}
\label{fig_scan_match_0095}
}
\subfloat
{
\begin{tikzpicture}
\begin{axis}[
 ybar,
    x=0.6cm,
    bar width=0.1cm,
    enlargelimits=0.15,
    width=0.24\textwidth,
     height=.20\textwidth,
    legend style={at={(0.50,0.99)}, anchor=north,font=\tiny, draw=none, row sep=-1ex},
    legend image post style={scale=0.3},
    label style={font=\scriptsize},
    x tick label style={font=\tiny},
    y tick label style={font=\scriptsize},
    y label style={at={(axis description cs:.25,.5)},rotate=0,anchor=south},
    ylabel={Time (sec)},
    symbolic x coords={baseline, 0xUp, 1xUp, 2xUp, 3xUp },
    scaled y ticks = {base 10:-2},
    xtick=data
    ]
\addplot coordinates {(baseline,75) (0xUp,28) (1xUp,40) (2xUp,112) (3xUp,296) };
\addplot coordinates {(baseline,67) (0xUp,27) (1xUp,33) (2xUp,79) (3xUp,251) };
\addplot coordinates {(baseline,28) (0xUp,7) (1xUp,11) (2xUp,46) (3xUp,339) };
\legend{GEN-ICP, STD-ICP, P2P-ICP}
\end{axis}
\end{tikzpicture}
\label{fig_timing_0095}
}
\subfloat
{
\begin{tikzpicture}
\begin{axis}[
 ybar,
    x=0.6cm,
    bar width=0.1cm,
    enlargelimits=0.15,
    width=0.24\textwidth,
     height=.2\textwidth,
    legend style={at={(0.65,0.99)}, anchor=north,font=\tiny, draw=none, row sep=-1ex},
    legend image post style={scale=0.3},
    label style={font=\scriptsize},
    x tick label style={font=\tiny},
    y tick label style={font=\scriptsize},
    y label style={at={(axis description cs:.25,.5)},rotate=0,anchor=south},
    ylabel={Map Error (m)},
    symbolic x coords={baseline, 0xUp, 1xUp, 2xUp, 3xUp },
    xtick=data
    ]
\addplot coordinates {(baseline,.10) (0xUp,.09) (1xUp,.05) (2xUp,.036) (3xUp,.010) };
\addplot coordinates {(baseline,1.07) (0xUp,.06) (1xUp,.01) (2xUp,.010) (3xUp,.030) };
\addplot coordinates {(baseline,1.08) (0xUp,.09) (1xUp,.037) (2xUp,.021) (3xUp,.040) };

\legend{GEN-ICP, STD-ICP, P2P-ICP}
\end{axis}
\end{tikzpicture}
\label{fig_map_error_0095}
}
\subfloat
{
\begin{tikzpicture}
\begin{axis}[
  ybar,
    x=0.6cm,
    bar width=0.1cm,
    enlargelimits=0.15,
    width=0.24\textwidth,
     height=.2\textwidth,
    legend style={at={(0.65,0.99)}, anchor=north,font=\tiny, draw=none, row sep=-1ex},
    legend image post style={scale=0.3},
    label style={font=\scriptsize},
    x tick label style={font=\tiny},
    y label style={at={(axis description cs:.2,.5)},rotate=0,anchor=south},
    y tick label style={font=\scriptsize},
    ylabel={Texture Error},
   symbolic x coords={baseline, 0xUp, 1xUp, 2xUp, 3xUp },
    xtick=data
    ]
\addplot coordinates {(baseline,158) (0xUp,57) (1xUp,31) (2xUp,28) (3xUp,58) };
\addplot coordinates {(baseline,174) (0xUp,45) (1xUp,13) (2xUp,8) (3xUp,19) };
\addplot coordinates {(baseline,114) (0xUp,54) (1xUp,16) (2xUp,12) (3xUp,20) };

\legend{GEN-ICP, STD-ICP, P2P-ICP}
\end{axis}
\end{tikzpicture}
\label{fig_texture_error_0095}
}
\vspace{0.1ex}
\setcounter{subfigure}{0}
\subfloat[Scan matching analysis]
{
\begin{tikzpicture}
\begin{axis}[
  ybar,
    x=0.6cm,
    bar width=0.1cm,
    enlarge y limits=0.80,
    enlarge x limits=0.15,
    width=0.24\textwidth,
     height=.20\textwidth,
    legend style={at={(0.65,0.99)}, anchor=north,font=\tiny, draw=none, row sep=-1ex},
    legend image post style={scale=0.3},
     y label style={at={(axis description cs:.30,.5)},rotate=0,anchor=south},
    label style={font=\scriptsize},
    x tick label style={font=\tiny},
    y tick label style={font=\scriptsize},
    ylabel={Average Error (m)},
    symbolic x coords={baseline, 0xUp, 1xUp, 2xUp, 3xUp },
    xtick=data,
    scaled y ticks=base 10:2
    ]
\addplot coordinates {(baseline,.12) (0xUp,.117) (1xUp,.110) (2xUp,.100) (3xUp,.091) };
\addplot coordinates {(baseline,.10) (0xUp,.122) (1xUp,.111) (2xUp,.100) (3xUp,.101) };
\addplot coordinates {(baseline,.11) (0xUp,.111) (1xUp,.103) (2xUp,.100) (3xUp,.092) };

\legend{GEN-ICP, STD-ICP, P2P-ICP}
\end{axis}
\end{tikzpicture}
\label{fig_scan_match_0001}
}
\subfloat[Timing performance]
{
\begin{tikzpicture}
\begin{axis}[
 ybar,
    x=0.6cm,
    bar width=0.1cm,
    enlargelimits=0.15,
    width=0.24\textwidth,
     height=.20\textwidth,
    legend style={at={(0.50,0.99)}, anchor=north,font=\tiny, draw=none, row sep=-1ex},
    legend image post style={scale=0.3},
    label style={font=\scriptsize},
    x tick label style={font=\tiny},
    y tick label style={font=\scriptsize},
    y label style={at={(axis description cs:.25,.5)},rotate=0,anchor=south},
    ylabel={Time (sec)},
    symbolic x coords={baseline, 0xUp, 1xUp, 2xUp, 3xUp },
    scaled y ticks = {base 10:-2},
    xtick=data
    ]
\addplot coordinates {(baseline,52) (0xUp,33) (1xUp,36) (2xUp,56) (3xUp,101) };
\addplot coordinates {(baseline,84) (0xUp,3) (1xUp,39) (2xUp,56) (3xUp,110) };
\addplot coordinates {(baseline,16) (0xUp,20) (1xUp,14) (2xUp,21) (3xUp,42) };
\legend{GEN-ICP, STD-ICP, P2P-ICP}
\end{axis}
\end{tikzpicture}
\label{fig_timing_0001}
}
\subfloat[Mapping error]
{
\begin{tikzpicture}
\begin{axis}[
 ybar,
    x=0.6cm,
    bar width=0.1cm,
    enlargelimits=0.15,
    width=0.24\textwidth,
     height=.2\textwidth,
    legend style={at={(0.65,0.99)}, anchor=north,font=\tiny, draw=none, row sep=-1ex},
    legend image post style={scale=0.3},
    label style={font=\scriptsize},
    x tick label style={font=\tiny},
    y tick label style={font=\scriptsize},
    y label style={at={(axis description cs:.25,.5)},rotate=0,anchor=south},
    ylabel={Map Error (m)},
    symbolic x coords={baseline, 0xUp, 1xUp, 2xUp, 3xUp },
    xtick=data
    ]
\addplot coordinates {(baseline,.052) (0xUp,.0147) (1xUp,.0140) (2xUp,.0147) (3xUp,.0146) };
\addplot coordinates {(baseline,.033) (0xUp,.0089) (1xUp,.0093) (2xUp,.0126) (3xUp,.0130) };
\addplot coordinates {(baseline,.037) (0xUp,.0106) (1xUp,.0120) (2xUp,.0139) (3xUp,.0144) };

\legend{GEN-ICP, STD-ICP, P2P-ICP}
\end{axis}
\end{tikzpicture}
\label{fig_map_error_0001}
}
\subfloat[Texture quality]
{
\begin{tikzpicture}
\begin{axis}[
  ybar,
    x=0.6cm,
    bar width=0.1cm,
    enlargelimits=0.15,
    width=0.24\textwidth,
     height=.2\textwidth,
    legend style={at={(0.65,0.99)}, anchor=north,font=\tiny, draw=none, row sep=-1ex},
    legend image post style={scale=0.3},
    label style={font=\scriptsize},
    x tick label style={font=\tiny},
    y label style={at={(axis description cs:.2,.5)},rotate=0,anchor=south},
    y tick label style={font=\scriptsize},
    ylabel={Texture Error },
   symbolic x coords={baseline, 0xUp, 1xUp, 2xUp, 3xUp },
    xtick=data
    ]
\addplot coordinates {(baseline,152) (0xUp,6) (1xUp,6) (2xUp,6) (3xUp,6) };
\addplot coordinates {(baseline,145) (0xUp,6) (1xUp,6) (2xUp,6) (3xUp,5) };
\addplot coordinates {(baseline,155) (0xUp,7) (1xUp,7) (2xUp,6) (3xUp,5) };

\legend{GEN-ICP, STD-ICP, P2P-ICP}
\end{axis}
\end{tikzpicture}
\label{fig_texture_error_0001}
}
\caption{Performance Analysis of the framework. Columns (a) scan matching performance analysis, (b) Timing analysis while scan matching, (c) $\mu_{ME}$ of map error, and (d) $\mu_{TE}$ of texture error. The first and second row corresponds to the sequence $0095$ and $0001$ respectively.}
\end{figure*}

In the sequence $0095$, the STD-ICP, P2P-ICP fails to converge and show higher average error\footnote{ of the order of $1$m  and is clipped from the fig. \ref{fig_scan_match_0095} to accommodate smaller values} (Fig. \ref{fig_scan_match_0095}), however, with \texttt{pose-refinement} both of them converges for the complete sequence which indicates the effect of the proposed method. For the sequence $0001$, the baselines and 0xUp have similar registration error which decreases as upsampling factor is increased (Fig. \ref{fig_scan_match_0001}). Averaged over all the scan matching techniques, the point-to-point error (m) for seq $0095$, decreases from $0.14$ (baseline) to $0.08$ (0xUp), $0.06$ (1xUp), $0.03$ (2xUp) $0.023$ (3xUp) and for seq $0001$, it decreases from $0.12$ (baseline) to $0.118$ (0xUp), $0.115$ (1xUp), $0.10$ (2xUp), $0.08$ (3xUp). Though, the difference between maximum and minimum registration errors is $\sim0.12m$ (seq $0095$) and $\sim0.03m$ (seq $0001$), the small improvement has large positive effects on the 3D surface quality of the generated maps (Fig. \ref{fig_reg_quality}). Hence, we argue that the upsampling leads to a better scan-matching performance and it also improves the quality of the texture map. 
%
Further, fig \ref{fig_timing_0095}, \ref{fig_timing_0001} shows the scan matching timing performance of the baselines and the \texttt{pose-refinement} for various upsampling rates. We average the time over the number of frames in the respective sequence and observe that the time consumed for alignment is directly proportional to the number of points. Further, the keyword ``real-time'' in this paper refers that all modules of the algorithm except the scan alignment process execute in real time on the given CPU. However, a GPU version of the scan alignment algorithms can be used to achieve overall real time performance. Since our intention in this paper has been to improve the mapping and texture quality, we have reported the timing analysis only for a CPU.
\subsection{Map and Texture Analysis}
\label{sub_map_texture_analysis}

Labeled Ground truth for textured maps doesn't exist; hence, we project the LIDAR points into the camera and generate ground truth for the color texture. Later, we compare the generated textured map against the accurate 3D data provided by the LIDAR and the generated color texture ground truth. In order to have a fair comparison, we manually remove moving object points both from ground truth and the textured map by using publicly available tool CloudCompare \cite{girardeau2015cloud}. Moreover, We set the \texttt{foveal~regions} to 
\begin{equation*}
\begin{aligned}
0m &\leq near~blind~zone < 3m, \\
3m &\leq white~zone < 15m, \\
15m &\leq far~blind~zone
\end{aligned}
\end{equation*}
\par
In Table \ref{tab_top1}, we have shown $ME$, $TE$ and $MTME$ for the textured maps generated using proposed framework and those generated using baselines\footnote{align the scans, obtain color texture by projecting LIDAR point to the camera and accumulate in a local frame $L$}. We also compare $ME$ for seq $0095$ with \cite{romanoni2015incremental}, \cite{romanoni2017mesh}. For this we rely only on their results provided in the paper due to unavailability of their code. As the previous works, \cite{romanoni2017mesh, romanoni2015incremental, yousif2014real} are bound to produce only qualitative analysis of their texturing process, we compare our results for $TE$ and $MTME$ against the baselines. The best and the worst results of proposed and baselines are shown in \textbf{\color{blue}blue} and \textbf{\color{red}red}. \\
\subsubsection{\underline{\textbf{Map Error}}}
\par
For the seq. $0095$, the proposed framework achieves the best $\mu_{ME}$ of $0.010m$ in E6 and outperforms \cite{romanoni2015incremental} by $88\%$, \cite{romanoni2017mesh} by $87\%$ as well as the baselines GEN-ICP by $90\%$, STD-ICP by $99\%$, and P2P-ICP by $99\%$. While for the seq. $0001$, the proposed framework achieves best $\mu_{ME}$ of $0.008m$ in E4. For this sequence, all of the E4-E7 have almost similar $\mu_{ME}$ which is not the case with seq. $0095$. It indicates that \texttt{pose-refinement} is dependent on the 3D information in LIDAR scans. Its effect is also indicated by $\mu_{ME}$ for seq $0095$  as GEN-ICP in E3 is trapped in local minima due to oversampling.
\\
\subsubsection{\underline{\textbf{Texture Error}}}
\begin{figure}[t]
\centering
\vspace{1ex}
\subfloat[]
{
\includegraphics[width=0.98\linewidth,height=20ex]{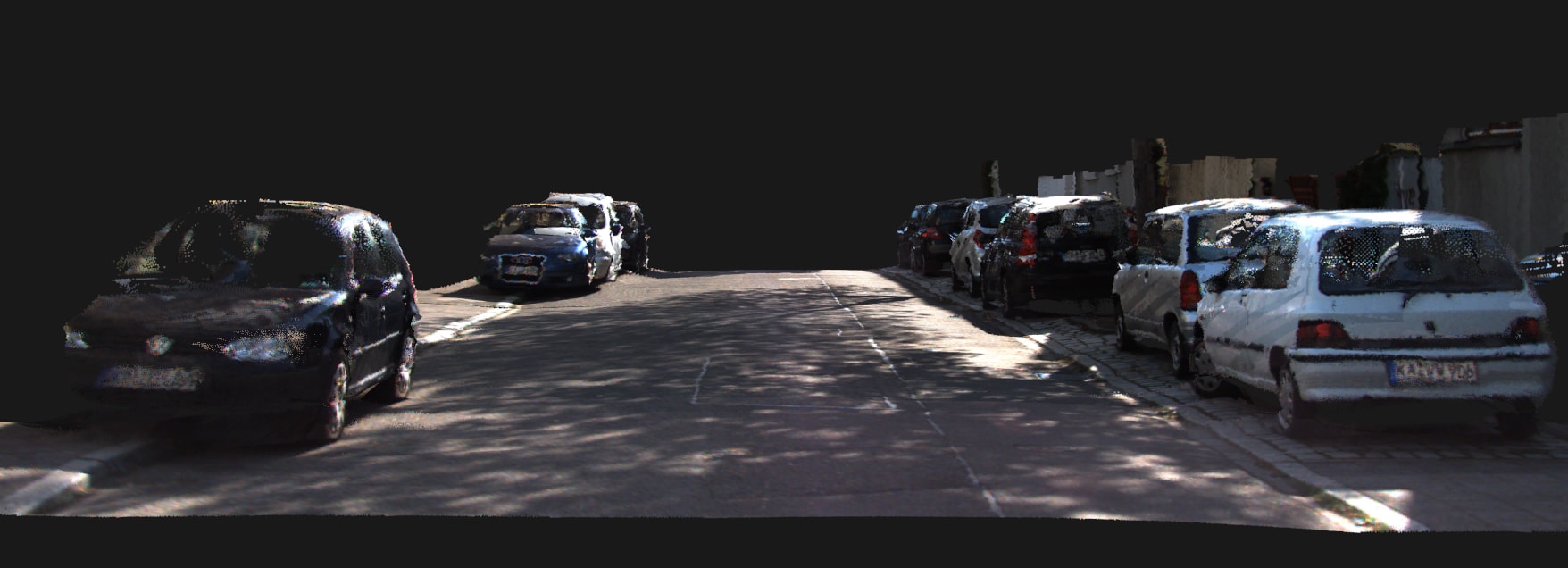}
}
\\
\subfloat[]
{
\includegraphics[width=0.98\linewidth,height=20ex]{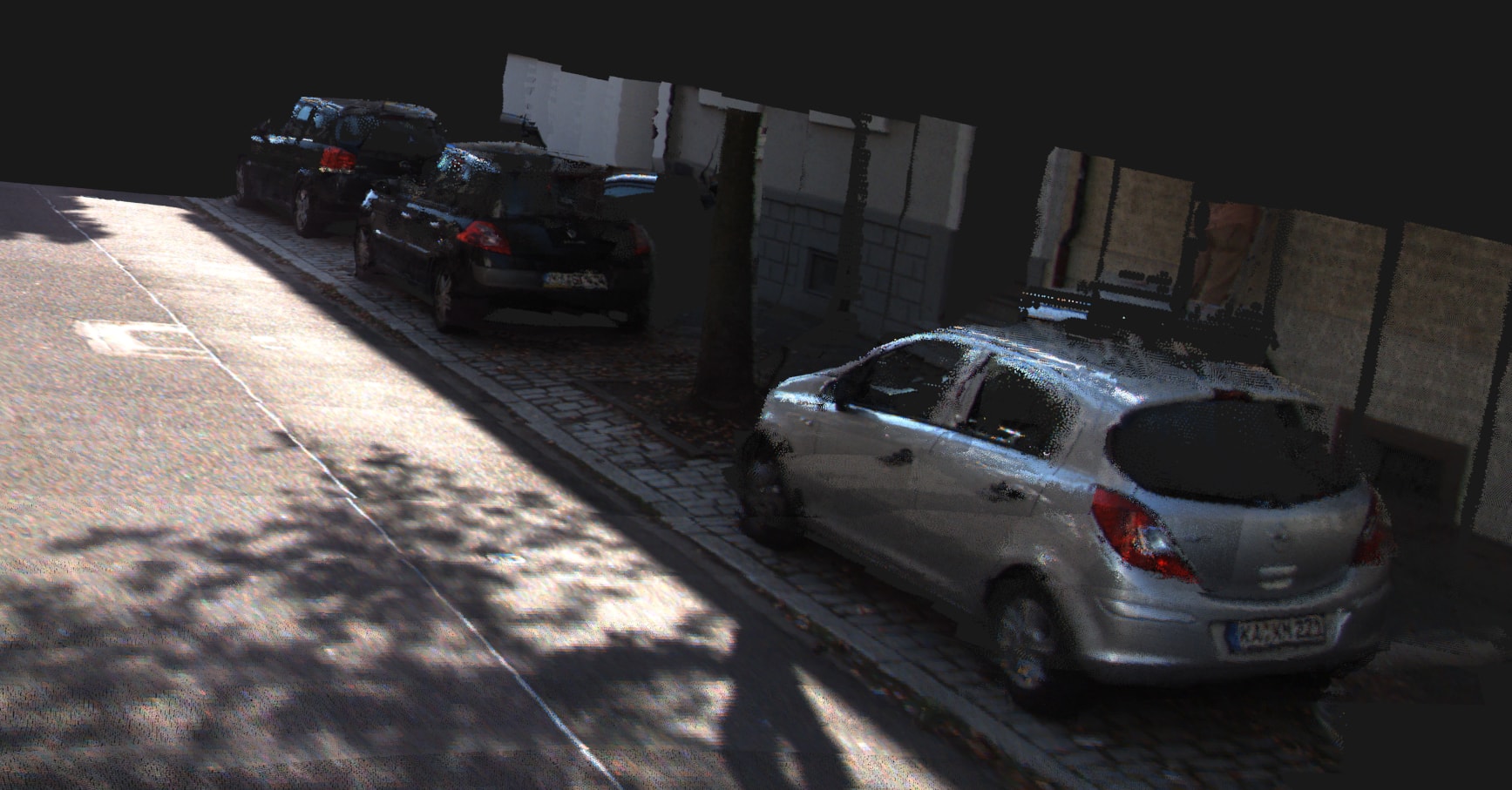}
}
\\
\subfloat[]
{
\includegraphics[width=0.98\linewidth,height=20ex]{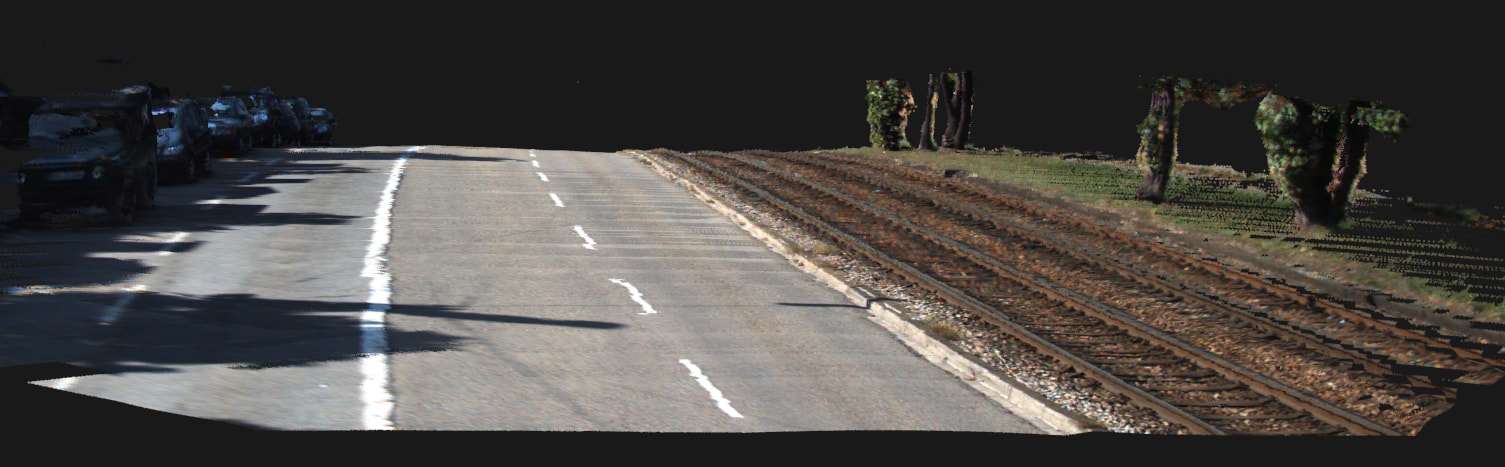}
}
\caption{Zoom in for better insight. Figure (a), (b) close up of the generated textured maps for sequence $0095$, and (c) sequence $0001$.}
\label{fig_texture_quality}
\end{figure}
Here, we claim that $ME$ doesn't convey any details about the texture quality of the map. It is indicated by large $TE$ in all of the baselines GEN-ICP, STD-ICP, P2P-ICP (Table \ref{tab_top1}). All of the baselines show $\mu_{TE}$ greater than $140$ which is $\sim54\%$ of the maximum attainable value $255$ for each of the \text{\color{red}red}, \text{\color{green}green}, \text{\color{blue}blue} component of a pixel color. This large error shows the usefulness of Eq. \ref{eq_texture_error}. 
\par
For both of the sequence, the proposed \texttt{ray-filtering} + \texttt{foveal-processing} technique drastically reduces the $TE$. It is evident from the $\mu_{TE}$ of E0-E11. The best achieved $\mu_{TE}$ for $0095$, $0001$ is $8.328$ and $5.24$ respectively which are only $\sim3\%$ and $2\%$ of the maximum value ($255$). These values of $\mu_{TE}$ are far superior to the best $145.85$ of the baselines without \texttt{ray-filtering} + \texttt{foveal-processing}. Effect of these improved $TE$ is clearly visible by realistic texture transfer as shown in fig. \ref{fig_texture_quality}. It should be noted that $TE$ also captures the effect of oversampling as seen in E3, E7 and E11, especially in the seq 0095.
\\
\subsubsection{\underline{\textbf{Mean Texture Mapping Error}}}
In order to asses the overall performance of the algorithm, we also report $MTME$ (Eq. \ref{eq_mtme}) for all of our experiments. From the table \ref{tab_top1}, especially from the baseline experiments, it can be seen that $MTME$ is high when either of $ME$ or $TE$ is high and attains lower values only when both of the $ME$ and $TE$ are lower. 
From the table \ref{tab_top1}, it can also be noticed that incorporation of \texttt{ray-filtering} reduces the $MTME$ drastically. For seq $0095$, E2 outperforms it's baseline version GEN-ICP by $95\%$, E6 outperforms STD-ICP by $98\%$ and E10 outperforms P2P-ICP by $98\%$. Similarly, for seq $0001$, all the three of E0-E11 outperforms their baselines by more than $95\%$. These visible effects of the reduced $MTME$ are shown in fig. \ref{fig_texture_quality}.
 \begin{table*}[!ht]  
\centering        
\captionsetup{justification=centering}
\caption{\\Quantitative evaluation of the proposed framework and various baselines. Here, ``pose-r'', ``ray-f'', ``fov-p'' stands for \texttt{pose-refinement, ray-filtering} and \texttt{foveal-processing} respectively.  }
\scriptsize
    \begin{tabular}{| c | c | c | c | c | c | c | c | c | c | c |}
    \hline
    \multirow{2}{*}{Algorithm} & \multicolumn{5}{c|}{KITTI Seq 0095} & \multicolumn{5}{c|}{KITTI Seq 0001} \\ \cline{2-11}    
                               & \multicolumn{2}{c|}{$ME$ (m)} & \multicolumn{2}{c|}{$TE$ } & \multirow{2}{*}{$MTME$} & \multicolumn{2}{c|}{$ME$ (m)} & \multicolumn{2}{c|}{$TE$ }  & \multirow{2}{*}{$MTME$} \\ \cline{2-5} \cline{7-10} 
      & $\mu_{ME}$ & $\sigma_{ME}$ & $\mu_{TE}$ & $\sigma_{TE}$  &  & $\mu_{ME}$ & $\sigma_{ME}$ & $\mu_{TE}$ & $\sigma_{TE}$ &  \\ \hline    

\cite{romanoni2015incremental}  & 0.089 & 0.131  & --  & -- & --  & -- & --  & -- & -- & --   \\ \hline
\cite{romanoni2017mesh}  & 0.082 & 0.098  & --  & -- & --  & -- & --  & -- & -- & -- \\ \hline

baseline GEN-ICP  & 0.102 & 0.089  & \text{\color{red}158.51}  & \text{\color{red}117.50}  & \text{\color{red}12.648} & \text{\color{red}0.052}  & \text{\color{red}0.064}  & \text{\color{red}152.66}  & \text{\color{red}90.51}  & \text{\color{red}7.744}\\ 
pose-r (GEN-ICP) + 0xUp + ray-f + fov-p (E0) & 0.098  & 0.089  & 57.42  & 74.60  & 6.899 & 0.014  & 0.045  & 6.49  & 22.97  & 0.098\\ 
pose-r (GEN-ICP) + 1xUp + ray-f + fov-p (E1) & 0.050 & 0.081 & 31.65  & 70.83  &2.300 & \text{\color{blue}0.014}  & \text{\color{blue}0.044}  & 6.37  & 22.61  & 0.094\\ 
pose-r (GEN-ICP) + 2xUp + ray-f + fov-p (E2) & \text{\color{blue}0.036}  & \text{\color{blue}0.075}  &  \text{\color{blue}28.39}  & \text{\color{blue}70.01}  & \text{\color{blue}0.513} & 0.014  & 0.045  & 6.13  & 21.93  & 0.095 \\ 
pose-r (GEN-ICP) + 3xUp + ray-f + fov-p (E3) & \text{\color{red}0.104}  & \text{\color{red}0.095}  & 58.78  & 87.81  & 7.643 & 0.014  & 0.045  & \text{\color{blue}6.11}  & \text{\color{blue}21.62}  & \text{\color{blue}0.092}\\ \hline

baseline STD-ICP  & \text{\color{red}1.076} & \text{\color{red}3.077}  & \text{\color{red}174.42}  & \text{\color{red}117.85}  & \text{\color{red}16.229} & \text{\color{red}0.033}  & \text{\color{red}0.046}  & \text{\color{red}145.85}  & \text{\color{red}90.93}  & \text{\color{red}4.835}\\ 
pose-r (STD-ICP) + 0xUp + ray-f + fov-p (E4) & 0.066  & 0.085  & 45.10  & 77.41  & 6.133 & \text{\color{blue}0.008}  & \text{\color{blue}0.033}  & 6.80  & 24.03  & \text{\color{blue}0.062}\\ 
pose-r (STD-ICP) + 1xUp + ray-f + fov-p (E5) & 0.010  & 0.040  & 13.17  & 49.44  & 0.286 & 0.009  & 0.032  & 6.77  & 23.38  & 0.064\\ 
pose-r (STD-ICP) + 2xUp + ray-f + fov-p (E6)  & \text{\color{blue}0.010}  & \text{\color{blue}0.042} & \text{\color{blue}8.328}  & \text{\color{blue}36.79}  & 0.268  & 0.012  & 0.041  & 6.27  & 22.35  & 0.078\\ 
pose-r (STD-ICP) + 3xUp + ray-f + fov-p (E7)  & 0.030  & 0.069 & 19.88  & 57.22  & 2.501 & 0.013  & 0.044  & \text{\color{blue}5.24}  & \text{\color{blue}20.36}  & 0.069\\ \hline

baseline P2P-ICP  &  \text{\color{red}1.086} &  \text{\color{red}3.084}  & \text{\color{red}170.32}  & \text{\color{red}114.47}  & \text{\color{red}15.814} & \text{\color{red}0.037}  & \text{\color{red}0.055}  & \text{\color{red}155.76} & \text{\color{red}90.17}  & \text{\color{red}5.759} \\ 
pose-r (P2P-ICP) + 0xUp + ray-f + fov-p (E8)  & 0.092  & 0.091  & 54.02  & 71.12  & 7.687 & \text{\color{blue}0.010}  & \text{\color{blue}0.038}  & 7.88  & 29.14  & \text{\color{blue}0.087} \\ 
pose-r (P2P-ICP) + 1xUp + ray-f + fov-p (E9) & 0.037  & 0.067 &  16.64  & 56.91  & 0.236 & 0.012  & 0.041  & 7.32  & 27.28  & 0.092\\ 
pose-r (P2P-ICP) + 2xUp + ray-f + fov-p (E10) & \text{\color{blue}0.021}  & \text{\color{blue}0.051}  &  \text{\color{blue}12.59}  &  \text{\color{blue}46.72}  & \text{\color{blue}0.201} & 0.013 & 0.045 & 6.25  & 23.59  & 0.092\\ 
pose-r (P2P-ICP) + 3xUp + ray-f + fov-p (E11) & 0.040 & 0.080  & 20.96  & 58.16  & 2.775 & 0.014 & 0.047  & \text{\color{blue}5.78} & \text{\color{blue}22.07} & 0.088 \\ \hline
   \end{tabular}
    \label{tab_top1}
\end{table*}

\section{Conclusion}
In this paper we presented robust framework to generate high quality textured 3D maps of urban areas. While development of this framework, we have focused on three major tasks: (\textit{i}) incremental accurate scan alignment, (\textit{ii}) real time dense upsampling of 3D scans and (\textit{iii}) color texture transfer without loosing fine grained details. The proposed framework successfully accomplishes all of the above three tasks by collectively leveraging multimodal information i.e. LIDAR scans, images, navigation data. The generated maps by using this framework, appears significantly realistic and carries fine grained details both in terms of 3D surface and color texture. 
Such high quality textured 3D maps can be used in several applications including precise localization of the vehicle, it can be used as a virtual 3D environment for testing various algorithms related to autonomous navigation without deploying the algorithm on a real vehicle and it can also be used as background map in computer games for real life gaming experience. 


\bibliographystyle{ieeetr}
\bibliography{main}

\appendices



\end{document}